\documentclass{article}
\usepackage{graphicx} 
\usepackage{multirow}%
\usepackage{amsmath,amssymb,amsfonts}%
\usepackage{amsthm}%
\usepackage{mathrsfs}%
\usepackage[title]{appendix}%
\usepackage{xcolor}%
\usepackage{textcomp}%
\usepackage{manyfoot}%
\usepackage{booktabs}%
\usepackage{longtable}%
\usepackage{algorithm}%
\usepackage{algorithmicx}%
\usepackage{algpseudocode}%
\usepackage{listings}%
\usepackage[numbers,sort&compress]{natbib}%
\usepackage{bibunits}
\usepackage{subcaption}%
\usepackage{authblk}
\usepackage{siunitx}
\usepackage{tabularx}
\usepackage{array}
\usepackage{pdflscape}
\usepackage{url}

\newcolumntype{L}[1]{>{\raggedright\arraybackslash}p{#1}}
\newcolumntype{Y}{>{\raggedright\arraybackslash}X}

\title{Wet-Biased Data Pipelines Mislead the Training and Evaluation of Generative Precipitation Nowcasting}
\author[1,*]{Christoph Metzl}
\author[1,2]{Stefan Kapsreiter}
\author[1]{Kianusch Vahid Yousefnia}
\author[3]{Virginia Poli}
\author[4]{Richard Müller}
\author[2]{Frank Krüger}
\author[1]{Tobias Bölle}

\affil[1]{Deutsches Zentrum für Luft- und Raumfahrt, Institut für Physik der Atmosphäre, Oberpfaffenhofen, Germany}
\affil[2]{Wismar University of Applied Sciences, Wismar, Germany}
\affil[3]{Arpae Emilia-Romagna, Hydro-Meteo-Climate Service (SIMC), Bologna, Italy} 
\affil[4]{German Weather Service,  Offenbach, Germany}
\date{}

\affil[*]{Correspondence and requests for materials should be
addressed to Christoph Metzl
(\texttt{christoph.metzl@dlr.de}).}

\begin{document}
\begin{bibunit}[naturemag]

\maketitle

\newpage

\section{Abstract}

Generative models for precipitation nowcasting are trained and evaluated predominantly on wet-biased datasets, which over-represent precipitation events to emphasize rare extremes. Selecting test cases based on target observations is known to favor forecasts that match the selected distribution rather than the intended deployment distribution. We show analytically how this target-dependent sampling also shifts the conditional distribution learned by generative models. When training and test sampling match, even an ideal model learns precisely the distribution favored by evaluation, allowing the evaluation bias to conceal the training bias. Using a fixed satellite-to-radar diffusion model, we demonstrate that target-dependent test sampling can reverse the calibration assessment. The model appears under-predictive on a wet-biased test set but over-predictive on a deployment-representative test set. This result is replicated outside the model's training domain after a seven-year gap. Utilizing classifier-free guidance as an example, we demonstrate how the contradictory calibration assessment can misdirect downstream model development.

\section{Introduction}

Extreme rainfall can endanger large populations through flooding \cite{tellman_satellite_2021} and cause substantial infrastructure damage \cite{koks_global_2019}. It is expected to become more frequent and intense in the face of climate change \cite{myhre_frequency_2019}, although the magnitude of these responses varies regionally \cite{pfahl_understanding_2017}. This underscores the need for timely and skillful forecasts of heavy precipitation \cite{majumdar_multiscale_2021}.

Nowcasting refers to weather forecasting at very short lead times, typically on the order of a few hours, with very high spatial and temporal resolution \cite{wang_guidelines_2017}. In this regime, numerical weather prediction (NWP) models face several limitations, including latency associated with observation processing and data assimilation, model spin-up, and insufficient spatiotemporal resolution \cite{sun_use_2014}. Although emerging Machine Learning (ML) based weather models achieve competitive forecast skill compared to NWP \cite{price_probabilistic_2025}, most still rely on NWP analyses for initialization and consequently inherit limitations associated with latency and resolution \cite{bouallegue_rise_2024}.

Traditional nowcasting methods, while data driven, are physically motivated and rely on motion field estimation and advection but struggle with forecasting the growth and decay of rain cells \cite{wilson_nowcasting_1998, pulkkinen_pysteps_2019, bolle_minimal_2026}. Early deterministic ML nowcasting models \cite{shi_convolutional_2015, ayzel_rainnet_2020} achieve superior aggregate skill relative to conventional extrapolation baselines, particularly for low precipitation, but their pixelwise regression objectives tend to blur aggressively, leading to overly smooth precipitation fields \cite{ayzel_rainnet_2020, franch_precipitation_2020}. Alternatively, nowcasting can be formulated as a probabilistic classification task over discrete intensity bins \cite{espeholt_deep_2022}. These probability maps can achieve strong verification scores, but representative deterministic fields may lack sharp, realistically organized precipitation structures \cite{agrawal_operational_2025}.

Therefore, generative modeling is an increasingly prominent direction in precipitation nowcasting \cite{ravuri_skilful_2021, zhang_skilful_2023, gao_prediff_2023, leinonen_latent_2023, gong_cascast_2024, yu_diffcast_2024, zhao_advancing_2024, foo_stldm_2025, ribeiro_flowcast_2026, wen_duocast_2026, ling_langprecip_2026, schusterbauer_probabilistic_2026}. Generative models like Generative Adversarial Networks (GANs)\cite{goodfellow_generative_2020}, diffusion \cite{song_scorebased_2021} and flow matching \cite{lipman_flow_2023} models aim to produce ensembles of spatially coherent precipitation fields.

Special emphasis is placed on forecast skill for extreme precipitation \cite{ravuri_skilful_2021, zhang_skilful_2023, nooten_improving_2023}, which is addressed through innovations in model architectures \cite{wang_skillful_2024, zhang_ms-edgecast_2026} and loss functions \cite{franch_gptcast_2025, gao_physics-guided_2026}. The rare event focus additionally motivates the evaluation on wet-biased test sets, which over-represent precipitation events relative to their climatological frequency. Wet-biased training sets also can help address the challenges of learning from rare events \cite{krawczyk_learning_2016}. Thus, a substantial number of studies and in particular seminal works \cite{veillette_sevir_2020, ravuri_skilful_2021, zhang_skilful_2023, gao_prediff_2023, gong_cascast_2024, yu_diffcast_2024} use wet-biased datasets for both training and testing. When cases are selected using the target observations, this practice shifts both the training and test distributions away from the intended deployment distribution. We refer to this mismatch as distribution shift.

The crucial risk is that evaluation can reward the very bias introduced during training. A generative model is trained to reproduce the rainfall probabilities represented in the sampled data. Evaluation on similarly sampled cases can therefore conceal their departure from the probabilities needed under deployment conditions. In practice, the assessment can even misidentify the direction of forecast bias: a model that over-predicts rain under deployment conditions may appear under-predictive on a wet-biased test set, suggesting an adjustment in the wrong direction. We examine this interaction theoretically and investigate its evaluation consequences through a controlled comparison of the same model on differently sampled test sets. 

On the evaluation side, verification on samples that have been selected based on the forecast targets or future observations is a known failure mode, referred to as the forecaster’s dilemma \cite{lerch_forecasters_2017}. This approach does not estimate performance over the intended deployment distribution anymore. Proper skill scores become improper \cite{gneiting_comparing_2011} and calibration diagnostics become misleading \cite{bellier_sample_2017}. It also breaks the covariate shift assumption that underpins a substantial part of the ML literature on counteracting distribution shift \cite{shimodaira_improving_2000, sugiyama_covariate_2007}. While the forecaster’s dilemma is well established, it has generally remained peripheral in nowcasting discussions up to now.

On the training side, the effect of target-dependent sampling on the training of generative nowcasting models is to our knowledge unexplored. Generative models learn to sample from the conditional distribution represented by their training set. Therefore, the distribution shift is directly built into the model itself.

As a concrete case we study these effects with a diffusion model that nowcasts radar reflectivity purely from geostationary satellite observations. While radar input is the gold standard for nowcasting, contiguous coverage is mostly limited to industrialized nations over land \cite{saltikoff_overview_2019}. The geostationary satellite, on the other hand, has almost global coverage and provides early detection of convective initiation \cite{schmetz_introduction_2002}. Thus, there is considerable recent interest in satellite to radar translation \cite{veillette_creating_2018, lee_validating_2024, qin_srdiff_2026, zeng_mcda-unet_2026} and satellite based severe weather nowcasting \cite{kucuk_transformer-based_2024, agrawal_operational_2025, dai_four-hour_2025, metzl_physical_2025, pathak_learning_2026}. Our model is trained over Germany on a wet-biased dataset and then evaluated on one wet-biased and one uniformly sampled test set, representative of the deployment distribution. Crucially, by fixing the model the effect of the evaluation distribution gets isolated. Facilitated through the satellite input choice, we repeat this experiment over Italy, which is fully outside of the model's training domain, seven years after the training period as an independent robustness test.

Finally, these findings are not an outright indictment of wet-biased datasets. They are useful for learning rare events, but distributional consequences need to be acknowledged. Specifically, evaluation has to be performed on the intended deployment distribution. We provide recommendations, and note the relevance to neighboring fields.

\section{Results}

\subsection{Sampling Practices in Generative Precipitation Nowcasting}\label{sec:audit}

Wet-biased training and test sets are common across the surveyed influential and recent generative precipitation nowcasting studies (Table \ref{tab:literature-sampling}). Additional work is reviewed in Supplementary Table \ref{tab:literature-sampling-additional}.

\begin{table*}[htbp]
    \centering
    \caption{Dataset sampling and calibration evaluation practices in a sample of influential and recent generative precipitation nowcasting studies. Additional work is surveyed in Supplementary Table \ref{tab:literature-sampling-additional}. Sampling implications are discussed in Section \ref{sec:audit}, calibration practices are surveyed in Section \ref{sec:forecaster_dilemma} and their implications discussed in Section \ref{sec:learningObjectiveTilt}. Deployment recovery indicates if the estimation of the verification scores on the deployment distribution by inverse probability weighting (IPW) is possible (cf. Section \ref{sec:forecaster_dilemma}).}
    \label{tab:literature-sampling}

    \small
    \setlength{\tabcolsep}{3pt}
    \renewcommand{\arraystretch}{1.12}

    \begin{tabularx}{\textwidth}{
        @{}
        L{1.75cm}
        L{1.35cm}
        Y
        Y
        L{1.85cm}
        L{2.45cm}
        @{}
    }
        \toprule
        Study &
        Model &
        Training sampling &
        Test sampling &
        Deployment recovery &
        Probabilistic calibration \\
        \midrule

        DGMR \cite{ravuri_skilful_2021} &
        GAN &
        Importance sampling &
        Stronger sampling &
        Possible &
        Reliability (IPW); rank histogram (shifted) \\

        \addlinespace
        NowcastNet \cite{zhang_skilful_2023}&
        GAN &
        Importance sampling &
        Stronger sampling and uniform sampling &
        Possible / uniform set &
        --- \\

        \addlinespace
        Netherlands DGMR \cite{nooten_improving_2023} &
        GAN &
        Deterministic filter &
        Equivalent to training &
        No &
        --- \\

        \addlinespace
        DiffCast \cite{yu_diffcast_2024}&
        Diffusion &
        Event-curated / deterministic filter &
        Equivalent to training &
        No &
        --- \\

        \addlinespace
        CasCast \cite{gong_cascast_2024}&
        Diffusion &
        Event-curated / deterministic filter &
        Equivalent to training &
        No &
        --- \\

        \addlinespace
        LDCast \cite{leinonen_latent_2023}&
        Diffusion &
        Intensity-balanced &
        Equivalent to training &
        Possible &
        Rank histogram (shifted) \\

        \addlinespace
        PDDN \cite{wang_skillful_2024}&
        Diffusion &
        Event-curated &
        Equivalent to training &
        No &
        --- \\

        \addlinespace
        Stormscope \cite{pathak_learning_2026}&
        Diffusion &
        Unclear &
        Uniform calibration set; strong- rain FSS set &
        Uniform set available &
        Reliability (uniform) \\

        \addlinespace
        FlowCast \cite{ribeiro_flowcast_2026} &
        Flow matching &
        Event-curated / uniform &
        Equivalent to training &
        Mixed &
        --- \\

        \addlinespace
        FREUD \cite{schusterbauer_probabilistic_2026}&
        Flow matching &
        Event-curated / uniform &
        Equivalent to training &
        Mixed &
        Rank histogram (shifted) \\

        \addlinespace
        GPTCast \cite{franch_gptcast_2025}&
        Autore- gressive &
        Deterministic filter &
        Event-curated &
        No &
        Rank histogram (shifted) \\

        \bottomrule
    \end{tabularx}
\end{table*}

We find that wet-biased training and test datasets are the norm in the literature reviewed on generative nowcasting. This result is consistent across all generative model types \cite{ravuri_skilful_2021, yu_diffcast_2024, franch_gptcast_2025, ribeiro_flowcast_2026} and applies both to meteorological and ML domains \cite{nooten_improving_2023, wang_skillful_2024, gong_cascast_2024, gao_prediff_2023}. 

Some works do not use the same sampling scheme for training and test sets \cite{ravuri_skilful_2021, zhang_skilful_2023, franch_gptcast_2025} motivating the three-distribution setup of Section \ref{sec:learningObjectiveTilt}.

The utilized sampling taxonomy differentiates between the following sampling categories: 'Importance sampling' refers to Kahn-style Monte Carlo importance sampling \cite{kahn_use_1955} which was first applied to precipitation nowcasting by \cite{ravuri_skilful_2021}. 'Deterministic filter' subsumes fixed inclusion rules such as minimum rainy-pixel fractions \cite{zhang_ms-edgecast_2026}, mean-rain thresholds \cite{yu_diffcast_2024} and rainy-day subsets \cite{gong_cascast_2024}. 'Intensity-balanced'  stratifies samples into bins based on precipitation intensity and then samples the bins equally \cite{leinonen_latent_2023}. 'Event-curated' datasets select samples on a case by case basis often focusing on heavy-rain \cite{wang_skillful_2024} or extreme \cite{franch_gptcast_2025} events. Specifically, we refer to the SEVIR dataset \cite{veillette_sevir_2020} as event-curated even though its storm events selected from the NOAA/NCEI Storm Events Database are complemented by a stochastic sampling component that is biased towards larger precipitation intensities. SEVIR is a highly influential benchmark dataset, which is used in a large number of nowcasting studies, including \cite{gao_prediff_2023, yu_diffcast_2024, gong_cascast_2024, zhao_advancing_2024, foo_stldm_2025, gao_physics-guided_2026, ribeiro_flowcast_2026, schusterbauer_probabilistic_2026, wen_duocast_2026}. This shows how a wet bias can be inherited silently by every downstream paper making use of a benchmark. 'Uniform' refers to unbiased sampling leading to a deployment consistent dataset. While comparatively rare, this strategy is employed by some works for a subset of their datasets either for both training and testing \cite{ribeiro_flowcast_2026, schusterbauer_probabilistic_2026} or only for restricted test scores \cite{zhang_skilful_2023, pathak_learning_2026}.

Importantly, all surveyed sampling schemes except for \cite{nooten_improving_2023} and the 'uniform' type make use of future observations to select samples and are thus target-dependent protocols.

\subsection{The Forecaster's Dilemma and Distribution Shift} \label{sec:forecaster_dilemma}

Selecting test cases using future observations changes which predictive distribution is rewarded by evaluation.

As established in the previous section, evaluating generative nowcasting models on wet-biased test datasets is a common practice. The wet bias is conventionally a product of target-dependent sampling i.e., the evaluation is conditioned on ground truth observations, $y$, because samples are selected retrospectively based on precipitation observed during the forecast period. However, forecasts should be evaluated conditionally only on information $x$ that is available at forecast time, not outcomes, $y$ \cite{gneiting_comparing_2011}. This failure mode is often termed the forecaster's dilemma \cite{lerch_forecasters_2017}: as forecasters are overwhelmingly judged on their performance on extreme events, it is tempting to restrict the evaluation solely to cases where extremes were indeed observed. Under these conditions it becomes beneficial to strongly over-predict extremes while skillful and calibrated forecasts are unjustly discredited whenever predictability is imperfect.

More precisely, a strictly proper scoring rule is defined as a verification score that is maximized by a forecast probability density $p(y\,|\,x)$ if and only if $p(y\,|\,x)$ reflects the forecaster's true beliefs \cite{winkler_good_1968}. Strictly proper scoring rules cannot be hedged or gamed, meaning by reporting a forecast probability density $p^*(y\,|\,x)$ deviating from ones true beliefs $p(y\,|\,x)$ about future events one cannot achieve a better score \cite{gneiting_strictly_2007}. Gneiting and Ranjan \cite{gneiting_comparing_2011} showed that restricting the test set by outcome is equivalent (up to scaling) to weighting a (strictly) proper score by some weight $w(y)$, which depends on the sampling strategy. The expected value of the score is then optimized by the shifted density
\begin{equation}
    p^*(y\,|\,x) \propto w(y)\,p(y\,|\,x)
    \label{eq:gneitingImproper}
\end{equation}
and not by the true predictive density $p(y\,|\,x)$. This constitutes a viable hedging strategy and thus renders strictly proper scoring rules improper. Crucially, $p^*(y\,|\,x)$ has exactly the mathematical form of the wet-biased training distribution that is simultaneously used to train the model (cf. Section \ref{sec:learningObjectiveTilt}). This problem can easily stay hidden as target-dependent sampling breaks calibration diagnostics like rank histograms as well \cite{bellier_sample_2017}. 

In ML terminology, evaluating generative nowcasting models on wet-biased test datasets, which introduces a mismatch between the test and deployment distribution, is a special case of distribution shift. However, the literature on standard ML distribution shift deals with covariate shift: the distribution of inputs, $p(x)$, differs between training and testing or deployment, while the conditional distribution of the target given the inputs, $p(y\,|\,x)$, remains unchanged \cite{shimodaira_improving_2000, sugiyama_covariate_2007}. This enables the study of post-hoc calibration methods, though they can fail in practice \cite{ovadia_can_2019, park_calibrated_2020}. Since most nowcasting dataset sampling schemes rely on target-dependent sampling, the conditional itself changes and the covariate shift assumption is violated. Post-processing on a wet-biased validation set therefore cannot restore deployment calibration even in principle.

Target-dependent sampling is also directly tied to the evaluation practices in the field. Most works only report raw pixel-based scores on wet-biased test sets. It is in theory possible to recover estimates of the verification scores on the actual deployment distribution by inverse probability weighting (IPW) with Horvitz-Thompson totals and H\'ajek-type ratios \cite{horvitz_generalization_1952}. To the best of our knowledge this has only been utilized by DGMR \cite{ravuri_skilful_2021} so far. Furthermore, IPW requires every candidate sample to have a nonzero chance of inclusion in the test set, which deterministic filter and event-curation approaches do not fulfill and thus excludes this possibility for a large set of works. In particular, it is not possible to apply IPW to the widespread SEVIR dataset \cite{veillette_sevir_2020} without major adjustments. A small subset of works report some pixel-based verification scores for a uniform sampling protocol \cite{zhang_skilful_2023, ribeiro_flowcast_2026, schusterbauer_probabilistic_2026}, which is faithful to a deployment scenario. Nevertheless, they often are presented as secondary. For example \cite{zhang_skilful_2023, schusterbauer_probabilistic_2026} do not present their uniform protocol scores in the main text and do not directly contrast them with the headline wet-biased scores. Therefore, potential misrepresentations of scores with respect to a deployment consistent evaluation that is predicted by the theory above, is not discussed. 

Moreover, calibration diagnostics are reported only rarely: Table \ref{tab:literature-sampling} contains all works that we are aware of that present any calibration measures, like rank histograms and reliability diagrams, at all. Among these, every published rank histogram is computed on a wet-biased test set. Deployment consistent calibration evidence exists only on the reliability side: DGMR \cite{ravuri_skilful_2021} reports IPW reliability diagrams, and Stormscope \cite{pathak_learning_2026} evaluates reliability on a uniformly sampled set. Notably, DGMR explicitly states that its rank histogram implementation cannot incorporate the importance weights \cite{ravuri_skilful_2021}. Consequently, even in the only work that corrects its verification scores, the rank histogram remains tied to the shifted test distribution. Critically, to the best of our knowledge there exists no controlled calibration comparison of a single generative nowcasting model evaluated on both a wet-biased and a deployment consistent test set. Closing this gap is a central contribution of this work.

\subsection{Target-Dependent Sampling Shifts the Learning Objective} \label{sec:learningObjectiveTilt}

Target-dependent training selection changes the conditional distribution that an ideal generative model learns.

\begin{figure}[htbp]
    \centering
    \includegraphics[width=\textwidth]{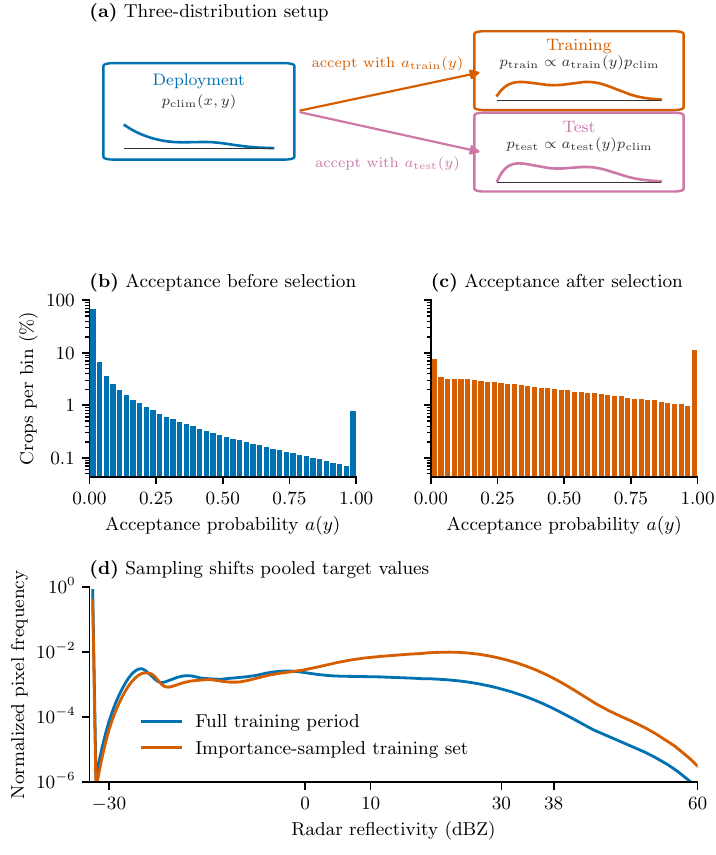}
    \caption{Importance sampling creates distinct deployment, training, and test distributions and shifts the empirical training data. \textbf{a}, The climatological deployment distribution $p_{\mathrm{clim}}(x,y)$ is shifted into training and test distributions by their respective target-dependent acceptance probabilities, $a_{\mathrm{train}}(y)$ and $a_{\mathrm{test}}(y)$. \textbf{b}, Distribution of the acceptance probability $a(y)$ over all candidate crops from the 2015--2018 training period. \textbf{c}, Distribution of the same acceptance probability after sampling, evaluated over the crops retained for model training.  Histograms in \textbf{b} and \textbf{c} show the percentage of crops per bin on a logarithmic scale. \textbf{d}, Normalized pooled radar-reflectivity distributions for all pixels in the complete training period and for the importance sampled crops actually used to train the model.}
    \label{fig:conceptual-sampling-shift}
\end{figure}

Let $x \in \mathbb{R}^{C \times T_x \times H \times W}$ denote a sequence of past observations serving as model input, where $C$ is the number of input channels, $T_x$ is the number of input time steps, and $H$ and $W$ are the height and width dimensions of the spatial grid dimensions respectively. The $C$ channels may contain any combination of fixed-grid data sources, such as radar composites, geostationary satellite imagery, or lightning observations. Let $y \in \mathbb{R}^{T_y \times H \times W}$ denote the corresponding future radar target sequence, consisting of $T_y$ forecast time steps on the same spatial grid. We treat both as realizations of the multivariate random variables $X$ and $Y$ with joint climatological density $p_{\mathrm{clim}}(x,y)$. This is our choice for the deployment distribution since it corresponds to a protocol assuming continuous operation. Though, this is not the only viable choice. The crucial point is that the deployment distribution can only be conditioned on information available at forecast time and cannot be conditioned on future observations. Hence, the commonly used target-dependent test datasets are ruled out. Therefore, an ideal generative nowcasting model (unlimited data, model capacity, and compute, with an optimizer that attains the true minimum) samples from the conditional density $p_{\mathrm{clim}}(y \,|\, x)$.

Dataset sampling is formalized by a binary selection variable $S \in \{0,1\}$ with acceptance probability $a(y)  \in [0,1]$:
\begin{equation}
    S \,|\, X = x, Y = y \;\sim\; \mathrm{Bernoulli}\bigl(a(y)\bigr),
    \label{eq:selection}
\end{equation}
where a candidate sample is included in the dataset if and only if $S = 1$. Although we borrow the term acceptance probability from Kahn-style importance sampling \cite{kahn_use_1955, ravuri_skilful_2021}, Eq.~\eqref{eq:selection} in principle applies to every sampling category of Section \ref{sec:audit} with a quantifiable inclusion rule. For example, deterministic filters correspond to the special case $a(y) \in \{0, 1\}$. Critically, $a$ depends on the target $y$, reflecting that all these schemes are target-dependent selection protocols. For some sampling schemes $a$ additionally depends on $x$. We omit this here as the central point is the dependence on $y$. Since training and test sets are sometimes constructed with different sampling schemes (cf. Section \ref{sec:audit}), the pipeline consists of three distinct distributions: the training distribution $p_{\mathrm{train}}(x,y) = p_{\mathrm{clim}}(x,y \,|\, S_{\mathrm{train}} = 1)$ with acceptance probability $a_{\mathrm{train}}(y)$, the test distribution $p_{\mathrm{test}}(x,y) = p_{\mathrm{clim}}(x,y \,|\, S_{\mathrm{test}} = 1)$ with acceptance probability $a_{\mathrm{test}}(y)$, and the deployment distribution $p_{\mathrm{clim}}(x,y)$ itself.

Applying Bayes' theorem to the conditional density under acceptance yields, for a generic acceptance probability $a_{\mathrm{train}}(y)$,
\begin{equation}
    p_{\mathrm{train}}(y \,|\, x)
    = \frac{\mathbb{P}(S_{\mathrm{train}}=1 \,|X=x, Y=y) \,p_{\mathrm{clim}}(y \,|\, x)}{\mathbb{P}(S_{\mathrm{train}}=1|\, X=x)}
    = \frac{a_{\mathrm{train}}(y)\,p_{\mathrm{clim}}(y \,|\, x)}{Z_{\mathrm{train}}(x)},
    \label{eq:conditional-tilt}
\end{equation}
where $\mathbb{P}$ denotes probability under the climatological distribution and the specified sampling mechanism. The normalizing factor
\begin{equation}
     Z_{\mathrm{train}}(x) = \mathbb{E}_{Y \sim p_{\mathrm{clim}}(\cdot \,|\, x)} \bigl[a_{\mathrm{train}}(Y) \bigr]
\end{equation}
is the expected training acceptance probability for a given input $x$, with $\mathbb{E}$ denoting expectation over target sequences $Y$ drawn from $p_{\mathrm{clim}}(y \,|\, x)$. The same derivation applies to the test distribution using its corresponding acceptance probability. Because $a(y)$ depends on $y$, the conditional density itself changes. This distinguishes target-dependent sampling from covariate shift, for which $a(y)$ would depend on $x$ only and Eq.~\eqref{eq:conditional-tilt} would reduce to $p_{\mathrm{train}}(y \,|\, x) = p_{\mathrm{clim}}(y \,|\, x)$. The sampling additionally shifts the input marginal, $p_{\mathrm{train}}(x) \propto Z_{\mathrm{train}}(x)\,p_{\mathrm{clim}}(x)$, leading to an over-representation of inputs, $x$, that precede rain. However, it is the shifted conditional that is learned by the model and that no covariate-shift correction can address.

Generative models learn to sample from the conditional distribution represented by their training set. An ideal generative model trained on a wet-biased dataset therefore reproduces $p_{\mathrm{train}}(y \,|\, x)$ of Eq.~\eqref{eq:conditional-tilt} rather than $p_{\mathrm{clim}}(y \,|\, x)$: the distribution shift is built directly into the model. Comparing with Equation~\eqref{eq:gneitingImproper} from Section \ref{sec:forecaster_dilemma}, Eq.~\eqref{eq:conditional-tilt} is exactly the hedging density $p^*(y \,|\, x) \propto w(y)\, p(y \,|\, x)$ of Gneiting and Ranjan \cite{gneiting_comparing_2011} with weight $w(y) = a_{\mathrm{train}}(y)$. An ideal model trained on the wet-biased training set thus represents the optimal hedge against evaluation on a test set constructed with the same acceptance probability. The test bias perfectly conceals the training bias when $a_{\mathrm{test}}(y) = a_{\mathrm{train}}(y)$. Consequently, a substantial number of works train the optimal cheater for their own improper benchmark and would deploy them on a distribution where the hedge is detrimental.

For score-based generative models the shift takes a particularly transparent form. Taking the logarithm and the gradient with respect to $y$ of Eq.~\eqref{eq:conditional-tilt}, the $x$-only normalizer $Z_{\mathrm{train}}(x)$ drops out and the learned score decomposes as
\begin{equation}
    \nabla_y \log p_{\mathrm{train}}(y \,|\, x)
    = \nabla_y \log p_{\mathrm{clim}}(y \,|\, x)
    + \nabla_y \log a_{\mathrm{train}}(y).
    \label{eq:score-tilt}
\end{equation}
An ideal score-based model hence learns the deployment score augmented by a deterministic drift up the gradient of the log acceptance probability. The wet bias is directly part of the training objective itself. Equation \eqref{eq:score-tilt} is exact at the data level. At positive diffusion noise levels, the statement holds in mollified form i.e. the shift is convolved with the noise kernel. 

The structure of the drift predicts where the wet bias should primarily materialize. For the Kahn-style importance sampling \cite{kahn_use_1955} used in this work and in \cite{ravuri_skilful_2021, zhang_skilful_2023} the acceptance probability for a sample $y$ is defined as
\begin{equation}
    a(y) = \min\left\{1, q_{\min} + \frac{m}{D} \sum_{j=1}^{D} \left[1-\exp\left(-\frac{y_j}{s}\right)\right] \right\}.
    \label{eq:acceptanceProb}
\end{equation}
Here, $D=T_yHW$ is the total number of pixel values across the target sequence, and $y_j$ denotes the rain rate at its $j$th entry. The parameter $q_{\min}$ sets the acceptance probability for an entirely dry sequence, $m$ controls the increase in acceptance probability with precipitation, and $s$ is a rain-rate scale controlling how quickly each pixel's contribution saturates. The specific parameter values chosen in this work are listed in Supplementary Table \ref{tab:importance-sampling-parameters}. The per-pixel derivative of $a(y)$ decays exponentially with intensity and vanishes entirely once the acceptance probability is clipped at one,
\begin{equation}
    \partial a/\partial y_j \propto e^{-y_j/s}.
\end{equation}
The drift therefore pushes hardest in the transition from a dry to light rain regime and barely at all at the extremes. This implies a stronger over-prediction of light rain, while high precipitation intensities are reinforced only marginally in comparison. We examine whether the observed forecast biases are consistent with this interpretation in Section \ref{sec:experiment1}.

Figure \ref{fig:conceptual-sampling-shift} illustrates the three distributions and the empirical sampling shift. Acceptance probabilities accumulate at the clipping limit, while rainy targets are enriched more strongly at light to moderate reflectivities than in the extreme right tail.

In addition, Eq.~\eqref{eq:conditional-tilt} is invertible. Thus, the deployment distribution can be recovered by reweighting with $1/a(y)$ and renormalizing. This corresponds to the IPW verification presented in Section \ref{sec:forecaster_dilemma} but one can also utilize IPW alternatively at the training or post-processing stages of the pipeline (Supplementary Section~\ref{sec:IPWTrainingAndPostporcessing}). As before, this requires $a(y) > 0$ for all candidate samples.

Furthermore, the discussion above holds in the idealized limit of unlimited data, model capacity, and compute, with an optimizer that attains the true minimum. In practice, rare extremes provide relatively few training examples, and a budget-constrained model may favor more frequent, lower-intensity precipitation patterns, leading to under-prediction of the strongest events even under wet-biased sampling and thus deviate from the predicted shift. The theory therefore establishes that the learning objective is misspecified. This does not mean that any trained model necessarily exhibits the full predicted bias, and it has to be demonstrated experimentally.

The limited and scattered published evidence compiled in Table \ref{tab:literature-sampling} is consistent with these predictions. All published rank histograms are computed on wet-biased test sets and appear skewed toward apparent under-prediction \cite{ravuri_skilful_2021, franch_gptcast_2025, schusterbauer_probabilistic_2026}, while the only two deployment consistent calibration diagnostics (the IPW reliability diagrams of DGMR \cite{ravuri_skilful_2021} and the uniform set reliability diagrams of Stormscope \cite{pathak_learning_2026}) both indicate over-prediction.  Though, the sampling scheme for the training set of Stormscope is not reported. DGMR presents both diagnostics side by side, an IPW diagram indicating over-prediction next to an unweighted rank histogram on the wet-biased test set suggesting mild under-prediction, but they do not comment on the apparent contradiction. It is exactly the sign flip in calibration that is predicted here. The nearly flat rank histogram of LDCast \cite{leinonen_latent_2023} is no exception. Its training and test sets share the same intensity-balanced shift, and by Eq.~\eqref{eq:conditional-tilt} a well-trained shifted model evaluated on its own shift should appear approximately calibrated. However, these pieces of evidence are uncontrolled. Models, sampling schemes, lead times, and diagnostics all differ across studies, and directions are judged visually from published figures, as none of these works report summary statistics of their histograms. In the following section we perform a comprehensive controlled comparison.

\subsection{Wet Bias Leads to Opposite Conclusions About the Same Model} \label{sec:experiment1}

We test experimentally whether target-dependent test sampling conceals forecast bias. Changing only the test-set sampling reverses the calibration diagnosis of the same model: the wet-biased set suggests under-prediction, while the uniform set establishes over-prediction.

We train an EDM-style diffusion model \cite{karras_elucidating_2022} to nowcast radar reflectivity solely from geostationary satellite imagery over Germany with a wet-biased training dataset. The model takes three satellite frames at $\SI{15}{\min}$ intervals ($t-30$, $t-15$, and $t$ minutes) as input and predicts radar reflectivity at $t$, $t+15$, and $t+30$ minutes. It is evaluated with $20$-member ensembles on two test sets from the same held-out period. One test set is uniformly sampled and thus representative of the deployment distribution, while  the other set is constructed with the same target-dependent Kahn-style importance sampling as the training set and is therefore wet-biased. Crucially, by fixing the model the effect of the evaluation distribution gets isolated as it is the only varied parameter. The identical frozen model is also evaluated with the same set up over Italy, which lies fully outside of the training domain, with a seven-year time gap as an independent robustness test.

\begin{table}[htbp]
    \centering
    \caption{Verification of the same nowcasting model on uniform and importance sampled (wet-biased) test sets over Germany and Italy. The IPW column re-estimates performance under the uniform sampling distribution from the importance sampled set using Horvitz-Thompson totals and H\'ajek-type ratios. Values in brackets are 95\% confidence intervals from a 10,000-replicate block bootstrap with seven-day blocks. Categorical scores and fractions skill score (FSS) are member averaged and aggregated over all lead times. Italian thresholds are the Marshall-Palmer equivalents of the German reflectivity thresholds (0.10, 2.61, and 9.55\,mm\,h$^{-1}$, respectively). Pooled CRPS is reported in dBZ for Germany and mm\,h$^{-1}$ for Italy and is therefore not comparable across domains.}
    \label{tab:test-set-score-comparison}
    \small
    \setlength{\tabcolsep}{3.5pt}
    \begin{tabular}{llccc}
        \toprule
        Metric & Threshold & \shortstack{Uniform\\test set}
        & \shortstack{Importance sampled\\test set}
        & \shortstack{IPW estimate from\\importance set} \\
        \midrule
        \multicolumn{5}{l}{\textit{Germany}} \\
        \addlinespace[2pt]
        \multirow{3}{*}{Bias}
        & 10\,dBZ & 1.541 [1.411, 1.705] & 0.688 [0.666, 0.715] & 1.540 [1.416, 1.697] \\
        & 30\,dBZ & 0.747 [0.643, 0.894] & 0.245 [0.209, 0.290] & 0.735 [0.635, 0.877] \\
        & 38\,dBZ & 0.687 [0.580, 0.856] & 0.233 [0.192, 0.287] & 0.667 [0.568, 0.820] \\
        \addlinespace
        \multirow{3}{*}{CSI}
        & 10\,dBZ & 0.198 [0.182, 0.213] & 0.363 [0.350, 0.376] & 0.198 [0.182, 0.213] \\
        & 30\,dBZ & 0.036 [0.028, 0.043] & 0.039 [0.032, 0.047] & 0.035 [0.028, 0.043] \\
        & 38\,dBZ & 0.017 [0.013, 0.022] & 0.015 [0.011, 0.019] & 0.017 [0.012, 0.021] \\
        \addlinespace
        \multirow{3}{*}{$\mathrm{FSS}_{128}$}
        & 10\,dBZ & 0.656 [0.628, 0.681] & 0.789 [0.779, 0.799] & 0.656 [0.630, 0.679] \\
        & 30\,dBZ & 0.273 [0.231, 0.321] & 0.237 [0.196, 0.290] & 0.267 [0.220, 0.322] \\
        & 38\,dBZ & 0.247 [0.188, 0.300] & 0.193 [0.143, 0.251] & 0.240 [0.178, 0.301] \\
        \addlinespace
        \shortstack{Pooled\\CRPS} & --- & 12.26 [12.00, 12.53] & 10.95 [10.70, 11.20] & 12.26 [11.98, 12.55] \\
        \midrule
        \multicolumn{5}{l}{\textit{Italy}} \\
        \addlinespace[2pt]
        \multirow{3}{*}{Bias}
        & 10\,dBZ & 1.453 [1.364, 1.574] & 0.675 [0.657, 0.692] & 1.447 [1.355, 1.561] \\
        & 30\,dBZ & 0.695 [0.635, 0.765] & 0.292 [0.262, 0.326] & 0.682 [0.625, 0.749] \\
        & 38\,dBZ & 0.881 [0.766, 1.016] & 0.360 [0.305, 0.421] & 0.848 [0.735, 0.977] \\
        \addlinespace
        \multirow{3}{*}{CSI}
        & 10\,dBZ & 0.190 [0.176, 0.201] & 0.356 [0.350, 0.363] & 0.191 [0.177, 0.203] \\
        & 30\,dBZ & 0.052 [0.047, 0.058] & 0.065 [0.059, 0.072] & 0.054 [0.048, 0.059] \\
        & 38\,dBZ & 0.020 [0.017, 0.024] & 0.022 [0.018, 0.025] & 0.020 [0.017, 0.023] \\
        \addlinespace
        \multirow{3}{*}{$\mathrm{FSS}_{128}$}
        & 10\,dBZ & 0.670 [0.646, 0.691] & 0.807 [0.798, 0.816] & 0.671 [0.646, 0.694] \\
        & 30\,dBZ & 0.332 [0.296, 0.369] & 0.313 [0.278, 0.348] & 0.337 [0.301, 0.372] \\
        & 38\,dBZ & 0.269 [0.221, 0.318] & 0.245 [0.198, 0.295] & 0.271 [0.226, 0.315] \\
        \addlinespace
        \shortstack{Pooled\\CRPS} & --- & 0.074 [0.060, 0.088] & 0.788 [0.710, 0.864] & 0.075 [0.061, 0.089] \\
        \bottomrule
    \end{tabular}
\end{table}

Table \ref{tab:test-set-score-comparison} presents pixel-based and categorical verification scores of the model for both datasets and both domains together with the IPW (Horvitz-Thompson / H\'ajek) estimates from the importance sampled sets. Complementary scores are listed in Supplementary Table~\ref{tab:complement-test-set-score-comparison}, and Supplementary Fig.~\ref{fig:rapsd-test-set-comparison} compares the radially averaged power spectra of forecasts and observations on both test sets. Square brackets indicate the $\SI{95}{\percent}$ confidence interval (CI). Some scores differ statistically significantly between the two German sets: CSI at $10$\,dBZ nearly doubles on the importance sampled set ($0.198$ $[0.182, 0.213]$ vs. $0.363$ $[0.350, 0.376]$), while the pooled CRPS is $\sim \SI{11}{\percent}$ better ($12.26$ $[12.00, 12.53]$ vs. $10.95$ $[10.70, 11.20]$). On top of the score inflation is the fact that the diagnosis changes qualitatively. On the importance sampled set the frequency bias is below $1$ at every threshold ($0.69/0.25/0.23$ at $10/30/38$\,dBZ). This implies a coherently  under-predictive model. On the uniform set the bias is $1.54$ at $10$\,dBZ but $0.75/0.69$ at $30/38$\,dBZ. Here over-prediction of light rain coexisting with under-prediction of moderate and strong rain. Therefore, the wet-biased set does not only change the score values, its directional conclusions are self-consistent but wrong.

\begin{figure}[htbp]
    \centering
    \includegraphics[width=\textwidth]{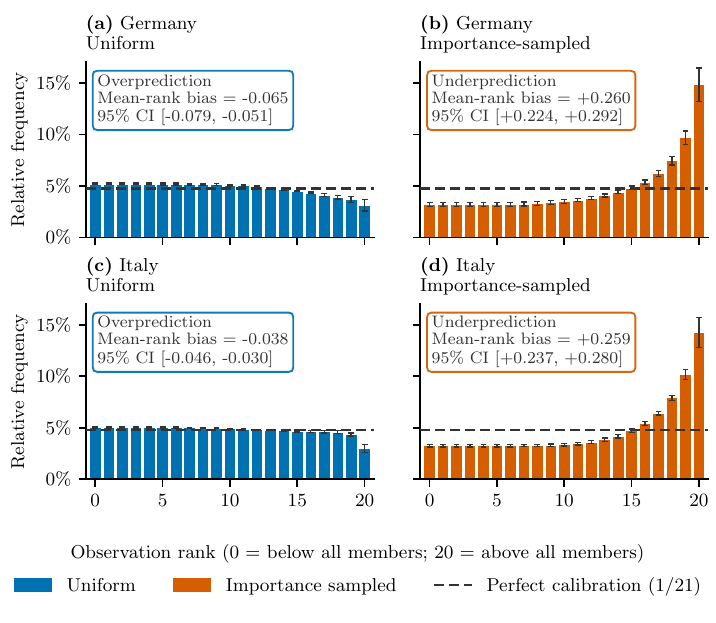}
    \caption{Rank histograms for the same 20-member forecast ensemble evaluated on uniform and importance sampled (wet-biased) test sets over Germany and Italy. Error bars and the reported mean-rank-bias intervals are \SI{95}{\percent} confidence intervals (CI) from a 10,000-replicate block bootstrap with seven-day blocks. The signed normalized mean-rank bias ranges from $-1$ (all observations below the ensemble) to $+1$ (all observations above the ensemble), with zero indicating a centered histogram. The dashed line denotes the relative frequency expected under perfect calibration. Histograms aggregate all lead times.}
    \label{fig:rank-histogram-test-set-flip}
\end{figure}

The rank histograms exhibited in Fig. \ref{fig:rank-histogram-test-set-flip} also demonstrate a sign flip in the calibration assessment of the model between the two test sets. The rank histogram skews to the right on the importance sampled set, indicating that the observations are above the ensemble too often, which implies an under-predictive model. However, on the uniform set the rank histogram skews to the left, signifying the observations are below the ensemble too often, which signals an over-predictive model. This result holds not just visually, but the mean-rank bias does indeed flip its sign statistically significantly between the uniform ($-0.065$ $[-0.079, -0.051]$) and the importance sampled ($+0.260$ $[+0.224, +0.292]$) test sets. Evaluating the same model one arrives at opposite calibration conclusions between the importance sampled and the uniform test set, which require opposite interventions (add rain vs. remove rain).

The sign flip demonstrates that target-dependent evaluation can conceal an existing forecast bias. This behavior is consistent with the two theoretical mechanisms discussed previously: (i) Rank histograms conditioned on observations can, misleadingly, skew towards under-prediction even for calibrated models \cite{bellier_sample_2017} (cf. Section~\ref{sec:forecaster_dilemma}). (ii) The shifted training dataset predicts a real wet bias (cf. Section \ref{sec:learningObjectiveTilt}). On the importance sampled test set (i) potentially masks the true bias of (ii), which is only observed on the uniform test set. As the model is kept fixed, the experiment isolates the effect of test-set selection, but it does not establish how much of the model’s light-rain over-prediction originates from training selection.

\begin{figure}[htbp]
    \centering
    \includegraphics[width=\textwidth]{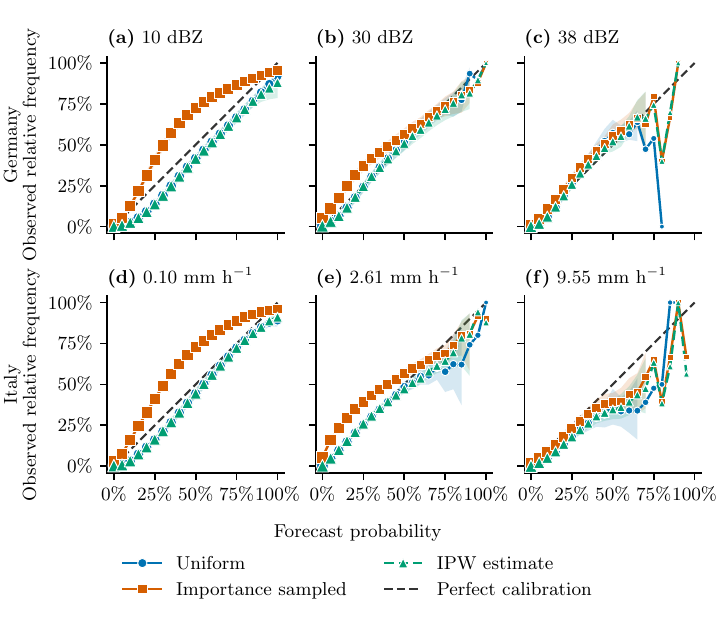}
    \caption{Reliability diagrams over Germany (top row) and Italy (bottom row) for threshold-matched exceedance levels. Columns show 10, 30, and 38\,dBZ over Germany and their Marshall-Palmer equivalents of 0.10, 2.61, and 9.55\,mm\,h$^{-1}$ over Italy. The same model is evaluated on the uniform test set, on the raw importance sampled (wet-biased) test set, and using an IPW (Horvitz-Thompson/H\'ajek) estimate from the importance sampled set. Shading denotes 95\% confidence intervals from a 10,000-replicate block bootstrap with seven-day blocks. Most intervals are narrower than the markers and are consequently obscured. An interval is omitted where fewer than ten seven-day blocks contribute forecasts to the corresponding probability bin. Marker area follows the logarithm of the corresponding refinement-bin count using one common scale across all panels and estimates. The diagonal denotes perfect reliability, and points aggregate all lead times.}
    \label{fig:reliability-germany-italy}
\end{figure}

The reliability diagrams displayed in Fig. \ref{fig:reliability-germany-italy} are fully consistent with this analysis. Changing the test set sampling protocol flips the calibration result at the $10$\,dBZ threshold, whereas the higher thresholds of $30$\,dBZ and $38$\,dBZ  remain comparatively close to perfect reliability over the supported bins.

The model's out of domain generalization works remarkably well. The CIs of most scores of the German and Italian domains presented in Table \ref{tab:test-set-score-comparison} intersect. Notably, the model even performs statistically significantly better on the Italian domain for the $30$\,dBZ threshold. The calibration diagnostics of Figs. \ref{fig:rank-histogram-test-set-flip} and \ref{fig:reliability-germany-italy} are in agreement as well. Crucially, the calibration sign flip observed over Germany is fully replicated over Italy. 

Furthermore, the IPW scores presented in Table \ref{tab:test-set-score-comparison}, which re-estimate performance under the uniform sampling distribution from the importance sampled set, fully agree with the uniform scores within error bars for both domains. The same is true for the IPW estimates of the reliability curves shown in Fig. \ref{fig:reliability-germany-italy}. This confirms that the mismatch between test sets is purely a sampling artifact, which can indeed be corrected through IPW under the condition of known, non zero acceptance probabilities.

Lastly, the observed over-prediction of light rain coexisting with under-prediction of moderate and strong rain indicates an imperfect model that behaves in a consistent manner with the interpretation from the previous section that the shift induced by the Kahn-style importance sampling is stronger at the transition from a dry to light rain regime than at the right tail of the distribution.

\subsection{How Wet-Biased Evaluation Steers Development Astray}
\label{sec:experiment2}

Increasing classifier-free guidance improves several forecast scores while amplifying over-prediction on the deployment consistent test set.

\begin{table}[htbp]
    \centering
    \caption{Verification of classifier-free guidance scales 1, 2, and 4 on the German uniform test set. Values in brackets are 95\% confidence intervals from a 10,000-replicate block bootstrap with seven-day blocks. Categorical scores and fractions skill score (FSS) are member averaged and aggregated over all lead times; CRPS is reported in dBZ. Guidance scale~1 is repeated from Table~\ref{tab:test-set-score-comparison} to make the guidance comparison self-contained.}
    \label{tab:guidance-score-comparison}
    \small
    \setlength{\tabcolsep}{3.5pt}
    \begin{tabular}{llccc}
        \toprule
        Metric & Threshold & Guidance scale 1 & Guidance scale 2 & Guidance scale 4 \\
        \midrule
        \multirow{3}{*}{Bias}
        & 10\,dBZ & 1.541 [1.411, 1.705] & 1.694 [1.563, 1.855] & 2.056 [1.887, 2.268] \\
        & 30\,dBZ & 0.747 [0.643, 0.894] & 1.685 [1.451, 2.020] & 3.761 [3.233, 4.527] \\
        & 38\,dBZ & 0.687 [0.580, 0.856] & 2.137 [1.804, 2.695] & 6.467 [5.354, 8.409] \\
        \addlinespace
        \multirow{3}{*}{CSI}
        & 10\,dBZ & 0.198 [0.182, 0.213] & 0.241 [0.222, 0.259] & 0.249 [0.227, 0.269] \\
        & 30\,dBZ & 0.036 [0.028, 0.043] & 0.064 [0.051, 0.076] & 0.067 [0.054, 0.080] \\
        & 38\,dBZ & 0.017 [0.013, 0.022] & 0.035 [0.026, 0.043] & 0.031 [0.023, 0.039] \\
        \addlinespace
        \multirow{3}{*}{$\mathrm{FSS}_{128}$}
        & 10\,dBZ & 0.656 [0.628, 0.681] & 0.714 [0.687, 0.738] & 0.709 [0.679, 0.736] \\
        & 30\,dBZ & 0.273 [0.231, 0.321] & 0.409 [0.345, 0.470] & 0.393 [0.336, 0.445] \\
        & 38\,dBZ & 0.247 [0.188, 0.300] & 0.400 [0.311, 0.464] & 0.265 [0.193, 0.326] \\
        \addlinespace
        \shortstack{Pooled\\CRPS} & --- & 12.26 [12.00, 12.53] & 8.34 [7.99, 8.69] & 6.80 [6.17, 7.44] \\
        \bottomrule
    \end{tabular}
\end{table}

\begin{figure}[htbp]
    \centering
    \includegraphics[width=\textwidth]{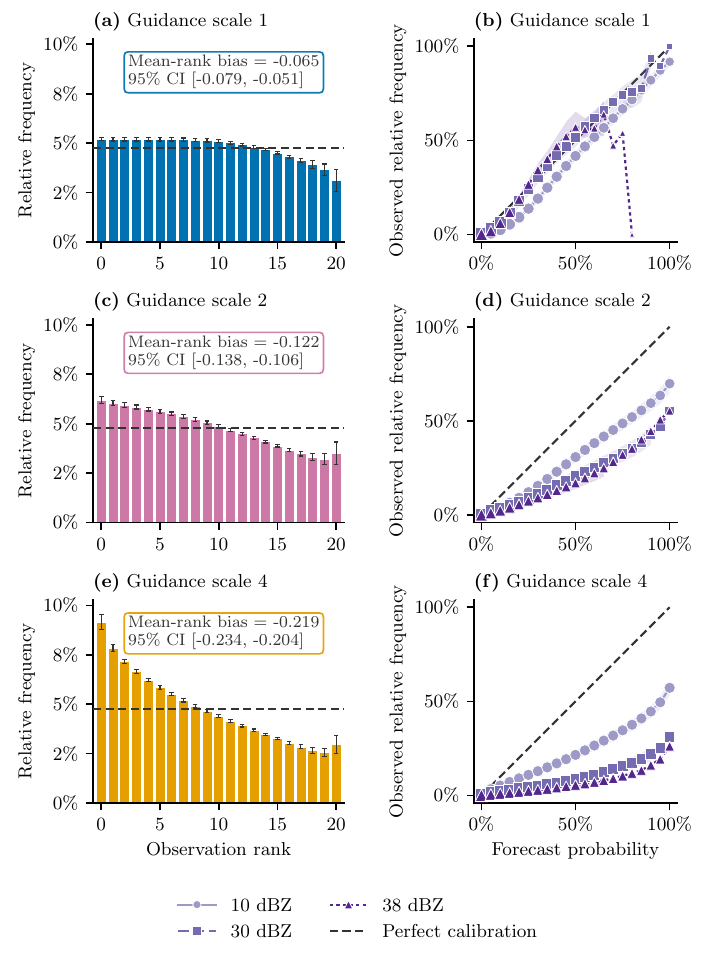}
    \caption{Calibration of classifier-free guidance scales 1, 2, and 4 on the German uniform test set. Each row shows one guidance scale. The left column contains rank histograms aggregated over all lead times. Error bars and confidence intervals are estimated the same way as in Fig. \ref{fig:rank-histogram-test-set-flip}. The right column contains reliability diagrams for exceedance of 10, 30, and 38\,dBZ. Confidence intervals and refinement-bin counts are depicted analogously to Fig. \ref{fig:reliability-germany-italy}. The dashed lines denote perfect calibration.}
    \label{fig:guidance-calibration}
\end{figure}

In the previous section we showed that on the importance sampled test set the model is diagnosed as under-predictive at all thresholds. This erroneous calibration assessment is a plausible motivation to indiscriminately increase the predicted rainfall amounts. We examine classifier-free guidance (CFG) \cite{ho_classifierfree_2021} as an intervention to achieve this in our experiments. Originally introduced as a sample-sharpening mechanism for image generation, CFG is increasingly used in generative precipitation nowcasting \cite{gong_cascast_2024,foo_stldm_2025,schusterbauer_probabilistic_2026} as it strengthens the influence of the conditioning information during sampling.

We evaluate guidance scales \(s\in\{1,2,4\}\) on the deployment consistent uniform German test set. Table \ref{tab:guidance-score-comparison} summarizes pixel-based scores; complementary scores are listed in Supplementary Table~\ref{tab:complement-guidance-score-comparison}. CSI at $38$\,dBZ doubles from $s=1$ to $s=2$ ($0.017$ $[0.013, 0.022]$ vs. $0.035$ $[0.026, 0.043]$), but levels off for $s=4$. Pooled CRPS on the other hand keeps improving with increasing guidance scale ($12.26$ $[12.00, 12.53]$ vs. $8.34$ $[7.99, 8.69]$ vs. $6.80$ $[6.17, 7.44]$). Evaluating solely aggregate scores commonly optimized in nowcasting would indicate that increasing $s$ appears highly beneficial. However, the calibration diagnostics presented in Fig. \ref{fig:guidance-calibration} reveal a substantial cost accompanying these score improvements. Increasing guidance progressively shifts observations toward low ensemble ranks and moves the reliability curves below the diagonal, demonstrating increasingly strong over-prediction. The frequency bias grows as well, especially at $38$\,dBZ, turning the previous under-prediction for $s=1$ ($0.687$ $[0.580, 0.856]$) into strong over-prediction at $s=4$ ($6.467$ $[5.354, 8.409]$). The improvement in pooled CRPS therefore coexists with worsening calibration and cannot, by itself, justify increasing guidance when reliable forecast probabilities are required.

\begin{figure}[htbp]
    \centering
    \includegraphics[width=\textwidth]{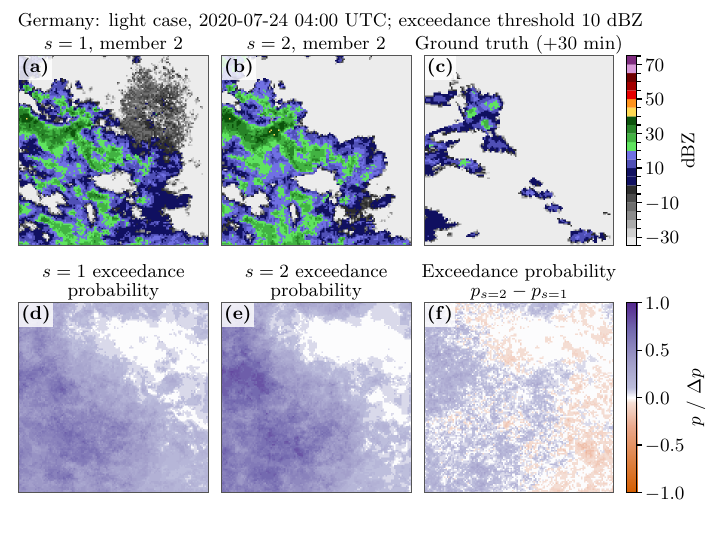}
    \caption{German light-rain case on 24 July 2020, 04:00 UTC and a lead time of 30 minutes. Each panel shows the same $128\times128$-pixel crop at $\SI{1}{\km}$ spatial resolution. \textbf{a,b}, The same paired ensemble member at guidance scales $s=1$ and $s=2$; \textbf{c}, ground truth. \textbf{d,e}, Exceedance probabilities $p_{s=1}$ and $p_{s=2}$, defined at each pixel as the fraction of the 20 ensemble members reaching or exceeding 10\,dBZ at the displayed lead time. \textbf{f}, The pixelwise probability difference $\Delta p=p_{s=2}-p_{s=1}$. Aggregated over all members, lead times, and pixels in this case, the observed exceedance fraction is 3.67\%, compared with forecast fractions of 23.58\% at $s=1$ and 26.87\% at $s=2$. The corresponding CSI changes only from 0.044 to 0.047, so the additional 3.29 percentage points of forecast event area yield little categorical benefit while strengthening the already excessive light-rain coverage.}
    \label{fig:case-germany-light}
\end{figure}

\begin{figure}[htbp]
    \centering
    \includegraphics[width=\textwidth]{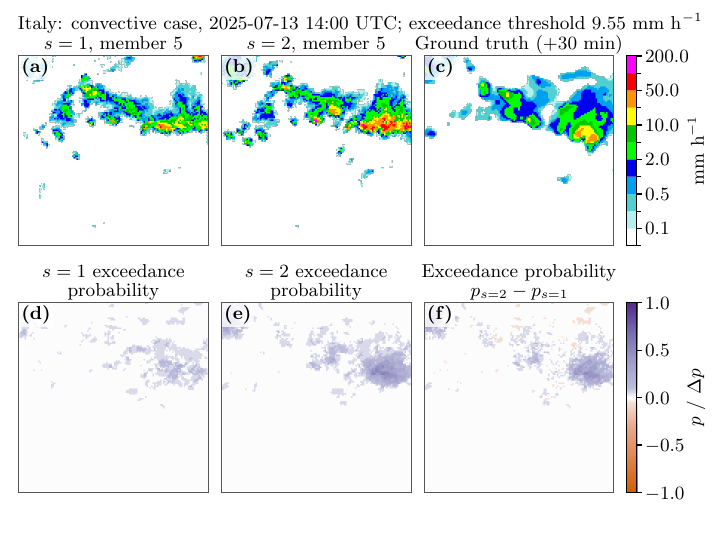}
    \caption{Italian convective case on 13 July 2025, 14:00 UTC and a lead time of 30 minutes. Each panel shows the same $128\times128$-pixel crop at $\SI{1}{\km}$ spatial resolution. \textbf{a,b}, The same paired ensemble member at guidance scales $s=1$ and $s=2$; \textbf{c}, ground truth. \textbf{d,e}, Exceedance probabilities $p_{s=1}$ and $p_{s=2}$, defined at each pixel as the fraction of the 20 ensemble members reaching or exceeding 9.55\,mm\,h$^{-1}$ at the displayed lead time. \textbf{f}, The pixelwise probability difference $\Delta p=p_{s=2}-p_{s=1}$. Aggregated over all members, lead times, and pixels in this case, increasing the guidance scale from $s=1$ to $s=2$ raises CSI from 0.057 to 0.168 and moves the forecast exceedance fraction from 0.55\% to 2.03\%, compared with 1.59\% in the observations. At the same time, the median member maximum increases from 94.67 to 322.04\,mm\,h$^{-1}$, illustrating that the improved localization and categorical overlap are accompanied by excessive peak intensities and additional forecast event area.}
    \label{fig:case-italy-convective}
\end{figure}

The case study exhibited in Fig. \ref{fig:case-germany-light} illustrates how the same intensity shift materializes in a light rain case. The case study exhibited in Fig. \ref{fig:case-italy-convective} illustrates this shift in a convective rain case. A German convective and an Italian light rain case together with the corresponding satellite inputs are shown in Supplementary Figs.~\ref{fig:case-germany-convective-inputs}--\ref{fig:case-italy-light}.

These findings largely agree with those of \cite{schusterbauer_probabilistic_2026}, who report that increasing CFG shifts forecasts toward stronger precipitation, initially improves most pixel-based scores, reduces ensemble spread, and eventually produces unrealistic intensities. The studies differ, however, in their CRPS response and in the conclusion drawn from this behavior. One possible contributor to the conflicting CRPS response is the estimator. While we evaluate the empirical forecast ensemble, the released evaluation pipeline of \cite{schusterbauer_probabilistic_2026} fits a Gaussian distribution from the ensemble mean and standard deviation. Because precipitation ensembles are generally non-Gaussian and CFG changes their shape and spread, these two CRPS formulations may result in different rankings of the guidance scales. Moreover, contrary to \cite{schusterbauer_probabilistic_2026} we consider CFG not intrinsically flawed. We argue that this depends on which distribution its scale is selected and for which forecast deficiency it is intended to compensate. The importance sampled test set incorrectly diagnoses our model as uniformly under-predictive and thereby supplies a plausible rationale for increasing guidance. Evaluation on the deployment consistent test set reveals that the same intervention amplifies an existing light-rain over-prediction and progressively destroys calibration. These results support selecting guidance on a deployment consistent validation dataset using both pixel-based scores and calibration diagnostics. They demonstrate a trade-off whose acceptability depends on the intended forecast use.

\section{Discussion}
\label{sec:discussion}

In this work we investigated the effects of wet-biased training and test sets caused by target-dependent sampling on generative precipitation nowcasting. We showed that target-dependent sampling applied to the training set shifts the learning objective of generative models. For an ideal model, this shift is equivalent to the optimal hedge against evaluation on a test set constructed with the same sampling protocol. These two shifts interact, causing the test bias to conceal the training bias. Agreement with the selected test distribution therefore does not establish calibration under deployment conditions.

This is not an outright indictment of importance sampling or wet-biased datasets. They are useful to reduce variance in budget constrained models as extremes need to be seen often enough during training to be learned. Our satellite to radar diffusion model underscores this fact. While it is over-predictive for light rain it still under-predicts strong rain when evaluated on a uniformly sampled dataset, even though its training set over samples it. The failure is leaving the induced distribution shift completely unacknowledged and evaluating the model on the same shifted distribution. As long as the sampling scheme's acceptance probabilities are known and strictly positive for every candidate sample this can be achieved through IPW, which can be applied to score reweighting (Section \ref{sec:forecaster_dilemma}) or in principle also at the training or post-processing stages (Supplementary Section \ref{sec:IPWTrainingAndPostporcessing}). 

To avoid inadequate use of wet-biased datasets we recommend the following:

\begin{enumerate}
    \item The intended deployment distribution should be stated explicitly and any differences between it and the training and test distribution should be flagged.
    \item Verification should be done on deployment consistent test sets. If importance sampling is utilized to save compute, acceptance probabilities should be recorded, sampling schemes that permit acceptance probabilities equal to zero for any candidate sample need to be avoided and reported scores should be IPW estimates. Alternatively forecast-based stratification can be used to emphasize extremes \cite{gneiting_comparing_2011, lerch_forecasters_2017}.
    \item Calibration measures like rank histograms and reliability diagrams ought to be reported to complement pixel-based scores.
    \item Post-processing should happen exclusively on deployment consistent validation datasets.
    \item In the longer term: The construction of a community benchmark for nowcasting with deployment consistent test sets analogously to WeatherBench for ML-based medium-range weather forecasting \cite{rasp_weatherbench_2020, rasp_weatherbench_2024}. 
\end{enumerate}
    
Several limitations remain: the theory presented only holds in the idealized limit and real models will always face some budget constraints. Extremes can remain under-predicted despite a wet-biased training set. Therefore, experimental confirmation is paramount.  We investigated one model family (EDM diffusion models) represented by one architecture, on one task (satellite to radar nowcasting), for one sampling scheme (Kahn-style importance sampling). Magnitudes will differ for radar to radar nowcasting, GANs, flow matching and different architecture and training algorithms. The mechanism itself, however, applies to generative objectives in general. A uniform test set is only a proxy for deployment. Further operational complications like data latency or quality effects are out of scope of this work. 

The mechanism we describe is not specific to nowcasting. It requires only a generative model fitted to a training set whose sample selection depends on the target, and a verification protocol that conditions the same way. This combination is likely to appear when rare events dominate the value of a model. For example, generative precipitation downscaling enriches its training data with rainy scenes in a similar way. Where the effect of this choice has been examined, it is consistent with our derivation \cite{harris_generative_2022}. ML climate emulation faces the same tension from the opposite side, as extremes are under sampled by the very simulations that supply the training data \cite{addison_machine_2022}, while the distribution of interest is a future climate. Our recommendations are mostly transferable: state the intended deployment distribution, retain the acceptance probabilities that separate it from the training and test distributions, and verify on the deployment distribution. Irrespective of the field, a skill claim is only as trustworthy as the distribution it was certified on.

\section{Methods}

\subsection{Data} \label{sec:data}

\begin{figure}[htbp]
    \centering
    \includegraphics[width=0.872\textwidth]{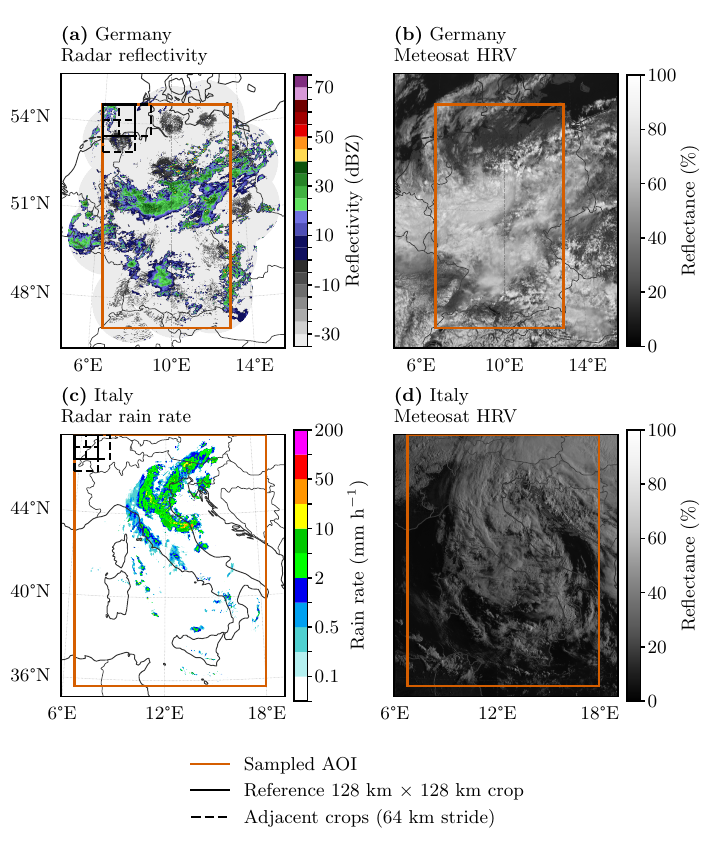}
    \caption{Geolocated input domains and crop construction. \textbf{a,b}, German RADOLAN WX radar reflectivity in dBZ and the corresponding Meteosat-10 SEVIRI high-resolution visible (HRV) channel at 1 June 2016, 11:45 UTC, near the peak of the Simbach flooding; \textbf{c,d}, Italian SRI radar rain rate in mm\,h$^{-1}$ and the corresponding HRV channel at 16 May 2023, 06:00 UTC, during the Emilia--Romagna flooding. Black outlines in the upper-left corner of each radar panel show one solid $128\times128$-pixel reference crop and its dashed right and lower neighbours.}
    \label{fig:domains-cropping}
\end{figure}

This section describes the data sources together with their processing and sampling routines for our satellite to radar nowcasting model for both the German and Italian domains, which are displayed in Fig. \ref{fig:domains-cropping} together with the cropping scheme. Additional details including software specifications can be found in Supplementary Section \ref{sec:data-processing-details}.

\paragraph{German radar.} The target over Germany is the DWD RADOLAN WX composite \cite{dwd_radolan}, a quality controlled national radar reflectivity mosaic in dBZ quantized in steps of 0.5\,dBZ. It is provided every 5\,min on a $900\times1100$ pixel grid with 1\,km spacing in the DWD polar-stereographic projection (WGS84 ellipsoid, central latitude $90^{\circ}$N, central longitude $10^{\circ}$E, true-scale latitude $60^{\circ}$N). We only use timestamps on the 15\,min grid to align the radar data with the lower temporal resolution of the satellite data. Pixels outside the composite coverage are flagged as missing by the product. We use the months May to October of the years 2015 to 2020 and split by calendar year: 2015--2018 for training, 2019 for validation, and 2020 for testing.

\paragraph{Italian radar.} The target over Italy is the IT-DPC-SRI surface rainfall intensity product \cite{franch_it-dpc-sri_2026} in mm\,h$^{-1}$. It is quality controlled and derived by the Italian Civil Protection Department from the national radar mosaic. The archive provides the product on a $1200\times1400$ pixel grid with 1\,km spacing in a transverse Mercator projection (WGS84 ellipsoid, latitude of origin $42^{\circ}$N, central meridian $12.5^{\circ}$E, scale factor 1, no false easting or northing). Since July 2020 the product is updated every 5\,min. We retrieve only the timestamps on the 15\,min grid from the continuously updated ArcoDataHub version of the archive. Pixels outside the composite coverage are flagged as missing by the product. We utilize the months May to October of 2025 as the Italian test period and process it analogously to the German data. The model was neither trained nor fine-tuned on Italian data.

\paragraph{Satellite imagery.} The input in both domains is Meteosat Second Generation (MSG) SEVIRI Level~1.5 imagery \cite{schmetz_introduction_2002} of the $0^{\circ}$ full-disc service with a 15\,min repeat cycle from the platforms Meteosat-10 in 2015--2017 and 2025 and Meteosat-11 in 2018--2020. Four of the 12 SEVIRI channels are used as predictors: the broadband 0.4–1.1\,$\mu$m high-resolution visible (HRV) and the 0.8\,$\mu$m visible (VIS008) channel as reflectance in percent, and the 10.8\,$\mu$m infrared (IR\_108) and 6.2\,$\mu$m water-vapor (WV\_062) channel as brightness temperature in K. No solar-angle correction is applied. The visible channels remain part of the input at night, where they are close to zero. Each scene is resampled with a nearest-neighbor approach onto the radar grid of the respective domain, so that satellite and radar fields share pixel coordinates. 

\paragraph{Cropping.} Samples are $128\times128$ pixel (128\,km) crops cut from a fixed lattice with a stride of 64 pixels, so that neighboring crops overlap by 50\%. Over Germany the lattice covers a window of $512\times896$ pixels of the full grid with vertices (NW, NE, SE, SW) at $(54.60^{\circ}\mathrm{N},5.85^{\circ}\mathrm{E})$, $(54.63^{\circ}\mathrm{N},13.56^{\circ}\mathrm{E})$, $(46.98^{\circ}\mathrm{N},12.88^{\circ}\mathrm{E})$, and $(46.96^{\circ}\mathrm{N},6.64^{\circ}\mathrm{E})$, producing $7\times13=91$ crops per timestep. Over Italy it covers $1024\times1344$ pixels with vertices at $(47.63^{\circ}\mathrm{N},5.49^{\circ}\mathrm{E})$, $(47.65^{\circ}\mathrm{N},19.12^{\circ}\mathrm{E})$, $(35.61^{\circ}\mathrm{N},17.98^{\circ}\mathrm{E})$, and $(35.60^{\circ}\mathrm{N},6.69^{\circ}\mathrm{E})$, producing $15\times20=300$ crops per timestep. The windows were chosen to minimize the number of pixels outside radar coverage.

\paragraph{Sample construction and quality control.} Every sample uses three satellite frames at $t-30$\,min, $t-15$\,min, and $t$ as input and three radar frames at $t$, $t+15$\,min, and $t+30$\,min as target. A timestamp is globally invalid if the radar file is missing, the satellite data is missing, or any of the twelve SEVIRI channels contains a missing value. Furthermore, a crop at a given timestamp is invalid if more than 360 of its 16,384 radar pixels are missing. This threshold was set to balance data quality against the number of retained crops over Germany. In total, these two rules mask 2.25\% of the German crop-timestamps. The same rules are applied unchanged over Italy to keep the two pipelines identical where possible. There, 61.8\% of all crop-timestamps are masked. This is due to many crops permanently lying outside the radar coverage. Finally, a six-frame sequence is considered only if all frames are valid for its crop and fall within May to October. The influence on sample count of each quality control mask is listed in Supplementary Table \ref{tab:qc-counts}. Missing radar values are filled before standardization. Over Germany, missing radar pixels are set to $-33$\,dBZ, just below the lowest product class of $-32.5$\,dBZ, and are thereby treated as no echo. Over Italy, missing rain rates are set to 0\,mm\,h$^{-1}$, i.e.\ they are treated as no rain. 

\paragraph{Importance sampling.} The German training set and the wet-biased test sets of both domains are constructed with the Kahn-style importance sampling of Section \ref{sec:learningObjectiveTilt}, which is adopted from DGMR \cite{ravuri_skilful_2021}. The acceptance probability $a(y)$ for a three-frame radar target sample $y$ in \,mm\,h$^{-1}$ is defined in Eq. \eqref{eq:acceptanceProb}. The specific parameter values chosen in this work are listed in Supplementary Table \ref{tab:importance-sampling-parameters}. Their sensitivity is analyzed in Supplementary Fig. \ref{fig:sampling-parameter-sensitivity} and Supplementary Table \ref{tab:sampling-rain-class-fractions}. Its effect on the training distribution is shown in Fig.~\ref{fig:conceptual-sampling-shift}. For Germany, the reflectivity target is converted to rain rate using the Marshall-Palmer relation $Z=256\,R^{1.42}$ before evaluating $a(y)$. For Italy, the rain rate is used directly. The same relation defines the Italian equivalents of the German verification thresholds (Section~\ref{sec:verification}). A candidate is accepted if a uniform random number does not exceed $a(y)$, using a fixed seed per split, and its acceptance probability is stored with the sample for IPW. In the German training period 425,713 of 6,221,943 candidates (6.84\%) are accepted. The training set is complemented by a uniformly sampled validation set of 2019 that is used for checkpoint selection (Section~\ref{sec:model}).

\paragraph{Uniform test sets and dataset sizes.} The deployment consistent test sets are uniform random samples without replacement from all candidate sequences of the test months. They are limited to the same sample count as the corresponding importance sampled set. Both German test sets contain 122,644 sequences from 2020 and both Italian test sets 95,111 sequences from 2025. Therefore, differences between the two protocols are not confounded by sample size. The importance sampled test sets use the identical acceptance function as the training set, i.e.\ $a_{\mathrm{test}}=a_{\mathrm{train}}$ in the notation of Section~\ref{sec:learningObjectiveTilt}. Supplementary Table~\ref{tab:dataset-sizes} summarizes all set sizes.

\subsection{Model} \label{sec:model}

This section specifies the normalization, diffusion formulation, the model architecture, the training algorithm, inference algorithm and the classifier-free guidance protocol. 

\paragraph{Normalization.} We utilize Z-score normalization, i.e all channels are standardized to zero mean and unit variance. Radar reflectivity uses the mean of $-3.9248$\,dBZ and standard deviation of $25.7472$\,dBZ of all target pixels of the importance sampled German training set. Each satellite channel uses the mean and standard deviation of the training period over the full German grid (Supplementary Table~\ref{tab:model-hyperparameters}). Over Italy, the satellite input is standardized with the German constants, so that the network always receives inputs on the scale it was trained on. Italian rain rates are stored as standardized $\log(1+R)$ for the ground truth only. They are transformed back to mm\,h$^{-1}$ before verification.

\paragraph{Diffusion formulation.} We follow the EDM formulation of Karras et al. \cite{karras_elucidating_2022}. Let $y$ denote the standardized three-frame radar target and $c$ the satellite conditioning. We wrap the network $F_\theta$ in the preconditioned denoiser
\begin{equation}
    D_\theta(\tilde y;\sigma,c) = c_{\mathrm{skip}}(\sigma)\tilde y + c_{\mathrm{out}}(\sigma) F_\theta\left(c_{\mathrm{in}}(\sigma)\tilde y; c_{\mathrm{noise}}(\sigma),c\right).
\label{eq:edm-precond}
\end{equation}
The preconditioning coefficients are
\begin{equation}
    c_{\mathrm{skip}}(\sigma) = \frac{\sigma_{\mathrm{data}}^2}{\sigma^2+\sigma_{\mathrm{data}}^2},
\label{eq:edm-cskip}
\end{equation}
\begin{equation}
    c_{\mathrm{out}}(\sigma) = \frac{\sigma\sigma_{\mathrm{data}}}{\sqrt{\sigma^2+\sigma_{\mathrm{data}}^2}},
\label{eq:edm-cout}
\end{equation}
\begin{equation}
    c_{\mathrm{in}}(\sigma) = \frac{1} {\sqrt{\sigma^2+\sigma_{\mathrm{data}}^2}},
\label{eq:edm-cin}
\end{equation}
\begin{equation}
    c_{\mathrm{noise}}(\sigma) = \frac{1}{4}\ln\sigma.
\label{eq:edm-cnoise}
\end{equation}
Only the noisy target is scaled by $c_{\mathrm{in}}$, while the conditioning $c$ is passed to the network without this scaling. Training minimizes
\begin{equation}
    \mathcal{L}(\theta) = \mathbb{E}_{(y,c),\,\sigma,\,n\sim\mathcal{N}(0,\sigma^2 I)}\Bigl[\lambda(\sigma)\,\bigl\lVert D_\theta(y+n;\sigma,c)-y\bigr\rVert_2^2\Bigr] ,
    \label{eq:edm-loss}
\end{equation}
with 
\begin{equation}
    \lambda(\sigma)=\frac{\sigma^2+\sigma_{\mathrm{data}}^2}{(\sigma\,\sigma_{\mathrm{data}})^2}.
\end{equation}
The squared error is averaged over all three target frames and all pixels. One noise level per sample is drawn from $\ln\sigma\sim\mathcal{N}(P_{\mathrm{mean}},P_{\mathrm{std}}^2)$. Supplementary Figure~\ref{fig:valueChain} summarizes one training step from the corruption of the target through the conditional denoiser to the loss. All parameters take the default values of \cite{karras_elucidating_2022} except for two changes induced by the unit variance of the standardized target. We set $\sigma_{\mathrm{data}}=1$ instead of $0.5$, and $P_{\mathrm{mean}}=-0.5$ instead of $-1.2$, which is the original noise-level prior shifted by $\ln 2$ so that the ratio of noise level to data scale covered during training is unchanged ($P_{\mathrm{std}}=1.2$ is retained). The score of the learned distribution follows from the denoiser as
\begin{equation}
    \nabla_{\tilde y}\log p(\tilde y;\sigma,c)=\bigl(D_\theta(\tilde y;\sigma,c)-\tilde y\bigr)/\sigma^2,
\end{equation}
which links the trained model to Eq.~\eqref{eq:score-tilt}. The IPW-trained model of Supplementary Section~\ref{sec:IPWTrainingAndPostporcessing} multiplies the per-sample loss by $1/a(y)$, normalized by the mean of this weight over the training set, and is otherwise identical.

\paragraph{Architecture.} $F_\theta$ is a conditional two-dimensional U-Net with $82.3\times10^6$ parameters that treats the three target frames as channels (Supplementary Figs.~\ref{fig:unet_architecture} and \ref{fig:diffusion_conditioning}). It operates on four resolution levels from $128\times128$ to $16\times16$ pixels with 128 to 1024 channels, two residual blocks per level, and one multi-head self-attention block in the bottleneck. The noise level enters every residual block through adaptive group normalization driven by a Fourier embedding of $c_{\mathrm{noise}}$. The satellite conditioning is processed by a separate encoder that applies two dimensional convolutions to each of the three input frames, adds a sinusoidal encoding of the time of day of each frame, and fuses the frames with a three dimensional convolution into 64 feature maps at full resolution. These features are concatenated with the noisy target at the network input and are additionally injected additively, after average pooling and a linear projection, at every coarser level of the encoder and decoder. Further architectural details are given in Supplementary Section \ref{sec:hyperparameters}.

\paragraph{Training.} The model is trained with AdamW with a batch size of 128 for at most 100 epochs with a five epoch linear warm-up followed by cosine decay of the learning rate. We apply gradient clipping, random rotations and flips of the crops, and an exponential moving average (EMA) of the weights. Checkpoint selection is based on the EDM loss of Eq. \eqref{eq:edm-loss} evaluated with the EMA weights on the uniformly sampled 2019 validation set. Early stopping terminates the training after ten epochs without improvement. The selected checkpoint is epoch 81. The training took about five days on a single NVIDIA HGX A100 80GB 500W GPU. Hyperparameters are listed in Supplementary Table~\ref{tab:model-hyperparameters}.

\paragraph{Inference and ensemble generation.} Forecasts are generated with the second-order multistep solver DPM-Solver++(2M) \cite{lu_dpm-solver_2025} in its data-prediction form with $N=50$ steps on the noise schedule of \cite{karras_elucidating_2022}
\begin{equation}
    \sigma_i=\bigl(\sigma_{\max}^{1/\rho}+\tfrac{i}{N-1}(\sigma_{\min}^{1/\rho}-\sigma_{\max}^{1/\rho})\bigr)^{\rho},
\end{equation} for $i=0,\dots,N-1$ with $\sigma_{\max}=80$, $\sigma_{\min}=0.002$, and $\rho=7$, followed by a final step to $\sigma=0$. Each of the 20 ensemble members starts from independent Gaussian noise $\mathcal{N}(0,\sigma_{\max}^2 I)$ whose seed is a deterministic function of the sample and member index. Hence, members with the same index start from identical noise across guidance scales and across the baseline and the IPW-trained model. This enables the paired comparisons shown in the case studies. The denoised output is de-standardized to dBZ without any clipping. Over Italy it is subsequently converted to mm\,h$^{-1}$ with the Marshall-Palmer relation.

\paragraph{Classifier-free guidance.} During training the satellite conditioning is dropped with probability 0.1, in which case the network denoises the target without any conditioning apart from the noise level (Supplementary Section~\ref{sec:hyperparameters}). At inference, classifier-free guidance \cite{ho_classifierfree_2021} with scale $s$ combines the conditional and the unconditional denoiser output as
\begin{equation}
    D_s(\tilde y;\sigma,c) = D_\theta(\tilde y;\sigma,\varnothing) + s\,\bigl[D_\theta(\tilde y;\sigma,c)-D_\theta(\tilde y;\sigma,\varnothing)\bigr]
    \label{eq:cfg}
\end{equation}
at every solver step, so that $s=1$ recovers the unguided model with one network evaluation per step and $s\neq1$ requires two. We evaluate $s\in\{1,2,4\}$. All results except the guidance study use $s=1$.

\subsection{Verification} \label{sec:verification}

This section presents our verification protocol, defines all verification metrics and our uncertainty estimation approach.

\paragraph{Protocol.} The model is evaluated with 20-member ensembles at lead times of 0, 15, and 30\,min on the uniform and the importance sampled test set of each domain. All reported scores aggregate the three lead times by pooling the underlying statistics (contingency counts, sums of squares, or sums of per-sample scores) before ratios are formed. Lead-time resolved scores are shown in Supplementary Fig. \ref{fig:skill-vs-lead-time}. Following DGMR \cite{ravuri_skilful_2021}, all scores except the fractions skill score (FSS) and the power spectra are computed on the central $64\times64$ pixels of each $128\times128$ crop to exclude boundary effects, whereas FSS and power spectra use the full crop because the largest neighborhood scale equals the crop size. Thresholds are 10, 30, and 38\,dBZ over Germany. Over Italy they are the equivalents 0.10, 2.61, and 9.55\,mm\,h$^{-1}$ obtained from the Marshall-Palmer relation $Z=256\,R^{1.42}$ and rounded to the product precision, so that categorical scores are comparable across domains up to the uncertainty of the Z--R relation. CRPS values are reported in the native unit of each domain and are not comparable across domains. We refer to \cite{Wilks_statistical_2019} for the definitions of all scores and to Supplementary Section~\ref{sec:verification-details} for implementation details. Scores omitted from the main tables, namely POD, FAR, FSS at 8 and 32 pixels, and the pixelwise CRPS, are listed in Supplementary Tables~\ref{tab:complement-test-set-score-comparison} and \ref{tab:complement-guidance-score-comparison}.

\paragraph{Categorical scores.} For each threshold, the 2 $\times$ 2 confusion matrix consists of the counts of true positives (TP), false positives (FP), false negatives (FN), and true negatives (TN), which are pooled over pixels, lead times, and samples. We report the probability of detection (POD), the false alarm ratio (FAR), the frequency bias (Bias), and the critical success index (CSI):
\begin{equation}
\begin{aligned}
\text{POD} &= \frac{\text{TP}}{\text{TP} + \text{FN}}, \quad
\text{FAR} = \frac{\text{FP}}{\text{FP} + \text{TP}}, \\
\text{Bias} &= \frac{\text{TP} + \text{FP}}{\text{TP} + \text{FN}}, \quad
\text{CSI} = \frac{\text{TP}}{\text{TP} + \text{FP} + \text{FN}}.
\end{aligned}
\end{equation}
Furthermore we report the FSS at a neighborhood size of $n\in\{8,32,128\}$ pixels. It compares the exceedance fractions $P^{f}$ and $P^{o}$ of forecast and observation obtained with an $n\times n$ moving average with zero padding,
\begin{equation}
    \mathrm{FSS}_n = 1-\frac{\sum_i \bigl(P^{f}_i-P^{o}_i\bigr)^2}{\sum_i \bigl(P^{f}_i\bigr)^2+\sum_i \bigl(P^{o}_i\bigr)^2},
    \label{eq:fss}
\end{equation}
where numerator and denominator are accumulated over all pixels, lead times, and samples before the ratio is taken. We compute these scores for every ensemble member separately and then average over members. This measures the skill of a typical member. We refrain from scoring the mean of a precipitation ensemble as it represents an overly smooth field and because averaging in the logarithmic dBZ unit is not equivalent to averaging rain rates.

\paragraph{Probabilistic scores.} The continuous ranked probability score (CRPS) is a proper scoring rule for the full predictive distribution. For an ensemble of $N=20$ members $x_1,\dots,x_N$ and the observation $x_a$ at one pixel we use its kernel representation \cite{gneiting_strictly_2007}
\begin{equation}
    \text{CRPS} = \frac{1}{N}\sum_{i=1}^{N}|x_i-x_a| - \frac{1}{2N^2} \sum_{i=1}^{N}\sum_{j=1}^{N}|x_i-x_j|,
    \label{eq:crps}
\end{equation}
which is the CRPS of the empirical cumulative distribution function of the ensemble. The pixel values are averaged over the pixels of a field, over all samples, and lead times. Because the pixelwise CRPS suffers from the double penalty problem, we additionally report a pooled CRPS similar to DGMR \cite{ravuri_skilful_2021}. The observation and every member are first smoothed with a $16\times16$ pixel average-pooling window with a stride of 4 pixels, and the CRPS is then computed on the pooled cells that cover the central domain.

\paragraph{Power spectra.} To assess whether the forecasts reproduce the spatial variability of the observations across scales, we compute radially averaged power spectral densities (RAPSD). Each $128\times128$ field is transformed with a two-dimensional discrete Fourier transform, the squared modulus is averaged over annuli of constant radial wavenumber, and the resulting one-dimensional spectra are averaged over ensemble members, samples, and lead times separately for forecasts and observations. The spectra of both domains and test sets are compared in Supplementary Fig.~\ref{fig:rapsd-test-set-comparison}. Since they are computed in the native unit of each domain, absolute levels are not comparable across domains.

\paragraph{Rank histograms.} The rank of the observation among the 20 sorted member values is computed for every pixel of the central domain, every lead time, and every sample leading to 21 bins. Pixels at which the observation and all members are below 10\,dBZ or 0.10\,mm\,h$^{-1}$ for the German and Italian domain respectively are excluded since their 21 values are all tied at no rain and thus their rank is arbitrary. Remaining values below that threshold are treated as tied and ties are broken uniformly at random. The histograms are summarized by the signed normalized mean-rank bias $b=2\,\bar{r}/M-1\in[-1,1]$, where $\bar r$ is the mean rank on the scale $0,\dots,M$ and $M=20$ is the ensemble size, so that $b<0$ indicates observations predominantly below the ensemble (over-prediction) and $b>0$ the opposite.

\paragraph{Reliability diagrams.} The forecast probability of threshold exceedance is the fraction of members exceeding the threshold at a pixel. Probabilities are binned into the 21 attainable values $k/20$, $k=0,\dots,20$, and the observed relative frequency of exceedance is computed per bin over the central domain, all lead times, and all samples. The refinement count per bin is indicated by the marker area. 

\paragraph{Inverse probability weighting.} Scores on the deployment distribution are re-estimated from the importance sampled test set by IPW with the recorded acceptance probabilities. Every sample $i$ receives the weight $w_i=1/a(y_i)$. Each statistic $S=\sum_i S_i$ accumulated over samples is replaced by its Horvitz-Thompson total \cite{horvitz_generalization_1952} $\hat S=\sum_i w_i S_i$, and scores that are ratios of two such statistics, i.e.\ all categorical scores, the FSS, and the reliability frequencies, become H\'ajek-type ratio estimators $\hat S_1/\hat S_2$. Per-sample scores are estimated by the self-normalized weighted mean $\sum_i w_i\,\mathrm{CRPS}_i/\sum_i w_i$. Rank histograms are not reweighted.

\paragraph{Uncertainty quantification.} Confidence intervals are obtained by a block bootstrap that respects the temporal correlation structure. Samples are assigned to fixed seven-day calendar blocks anchored at a common epoch, which yields 27 blocks per domain and test year. For each of 10,000 replicates, blocks are drawn with replacement, the per-block sufficient statistics are summed, and all scores, histograms, and reliability curves are recomputed from the summed statistics. Reported intervals are the 2.5th and 97.5th percentiles of the replicates. Comparisons between models or guidance scales use the same block draws for both runs (paired resampling).

\subsection{Use of AI-Assisted Tools} 
The authors used OpenAI Codex and Anthropic Claude Code to assist with software development and with drafting, revising and editing the manuscript. The research questions and scientific ideas originated with the authors. All AI-assisted outputs were reviewed and checked by the authors, who take full responsibility for the code, analyses and manuscript.

\section*{Data availability}
The RADOLAN WX radar composites were obtained from the Deutscher Wetterdienst (DWD) upon request. Enquiries regarding access to these historical composites should be directed to DWD’s data service (datenservice@dwd.de). Availability and access conditions are determined by DWD.
The MSG SEVIRI Level~1.5 satellite observations are available from the EUMETSAT Data Store (\url{https://data.eumetsat.int/}), subject to EUMETSAT’s access and licensing conditions. EUMETSAT has confirmed that these observations may not be redistributed in the processed form used in our machine-learning datasets. Consequently, the prepared paired satellite–radar datasets cannot be redistributed by the authors.
The Italian IT-DPC-SRI radar data are publicly available through ArcoDataHub under a CC BY-SA 4.0 license (\url{https://arcodatahub.com/datasets/notebooks/italian-radar-dpc-sri/}).

\section*{Code availability}
The source code for data processing, dataset sampling, model training, forecast generation and verification for both study domains is provided confidentially to editors and reviewers in the ZIP archive accompanying this submission. Public release of the code on GitHub is planned for publication, subject to institutional clearance; the present archive is provided for editorial assessment and peer review.

\putbib[references]

\section*{Acknowledgements}
C. M. carried out his contributions within the Italia–Deutschland Science-4–Services Network in Weather and Climate (IDEAS4S; SESTO, 4823IDEAP4). This Italian–German research network of universities, research institutes, and DWD was funded by the Federal Ministry of Digital and Transport (BMDV). We gratefully acknowledge the computational and data resources provided through the joint high performance data analytics (HPDA) project “terrabyte” of the DLR and the Leibniz Supercomputing Center (LRZ).

\section*{Author contributions}
C.M. conceived the study, developed the methodology and theoretical analysis, and supervised S.K.’s research. K.V.Y. contributed to the theoretical analysis. S.K. implemented the initial data processing pipeline and DDPM-based model. C.M. developed the final data processing and verification pipelines and EDM-based model, conducted the experiments, analysed the results, and wrote the original manuscript draft. V.P. contributed to radar data processing. T.B., K.V.Y., V.P. and R.M. supervised C.M.’s research. F.K. supervised S.K.’s research. T.B. acquired funding. All authors contributed to refining the research concept and reviewed and edited the manuscript.

\section*{Competing interests}
The authors declare no competing interests.

\end{bibunit}

\clearpage

\begin{bibunit}[naturemag]

\setcounter{figure}{0}
\setcounter{table}{0}
\setcounter{equation}{0}
\setcounter{subsection}{0}

\renewcommand{\figurename}{Supplementary Fig.}
\renewcommand{\tablename}{Supplementary Table}
\renewcommand{\thefigure}{\arabic{figure}}
\renewcommand{\thetable}{\arabic{table}}
\renewcommand{\theequation}{\arabic{equation}}
\renewcommand{\thesubsection}{\arabic{subsection}}
\renewcommand{\refname}{Supplementary References}

\begin{center}
    {\LARGE\bfseries Wet-Biased Data Pipelines Mislead the Training and
    Evaluation of Generative Precipitation Nowcasting\par}
    \bigskip
    {\Large Supplementary Information\par}
\end{center}
\bigskip


\subsection{Additional Sampling Practices Survey}

This section surveys the dataset sampling and calibration evaluation practices of additional generative precipitation nowcasting studies in Supplementary Table \ref{tab:literature-sampling-additional}, complementing Table~\ref{tab:literature-sampling}.

\begingroup
\small
\setlength{\tabcolsep}{3pt}
\renewcommand{\arraystretch}{1.12}

\setlength{\LTleft}{0pt}
\setlength{\LTright}{0pt}
\setlength{\LTcapwidth}{\textwidth}

\newlength{\literaturemiddlecol}
\setlength{\literaturemiddlecol}{%
    \dimexpr(\textwidth-7.40cm-10\tabcolsep)/2\relax
}

\begin{longtable}{
    @{}
    L{2.1cm}
    L{1.35cm}
    L{\literaturemiddlecol}
    L{\literaturemiddlecol}
    L{1.85cm}
    L{2.10cm}
    @{}
}
    \caption{Dataset sampling and calibration evaluation practices in additional generative precipitation nowcasting studies. Studies included in Table~\ref{tab:literature-sampling} are omitted.}
    \label{tab:literature-sampling-additional} \\

    \toprule
    Study &
    Model &
    Training sampling &
    Test sampling &
    Deployment recovery &
    Probabilistic calibration \\
    \midrule
    \endfirsthead

    \multicolumn{6}{@{}l}{%
        \tablename~\thetable\ continued
    }\\
    \toprule
    Study &
    Model &
    Training sampling &
    Test sampling &
    Deployment recovery &
    Probabilistic calibration \\
    \midrule
    \endhead

    \midrule
    \multicolumn{6}{r@{}}{\footnotesize Continued on next page} \\
    \endfoot

    \bottomrule
    \endlastfoot

    LangPrecip \cite{ling_langprecip_2026} &
    Flow matching &
    Event-curated; filtering unclear &
    Equivalent to training &
    Unclear &
    --- \\

    \addlinespace
    PostCast \cite{gong2025postcast} &
    Diffusion &
    Mixed; dataset-inherited &
    Random 500 per dataset from test splits &
    Mixed &
    --- \\

    \addlinespace
    PreDiff \cite{gao_prediff_2023} &
    Diffusion &
    Event-curated &
    Equivalent to training &
    No &
    --- \\

    \addlinespace
    DYffCast \cite{seal_dyffcast_2024} &
    Diffusion &
    Unclear &
    Unclear &
    Unclear &
    Spread--skill only \\

    \addlinespace
    DuoCast \cite{wen_duocast_2026} &
    Diffusion &
    Event-curated / deterministic filtering &
    Equivalent to training &
    No &
    --- \\

    \addlinespace
    MoCast+ \cite{wu_mocast_2026} &
    Diffusion &
    Event-curated / deterministic filtering &
    Equivalent to training &
    No &
    --- \\

    \addlinespace
    Nowcast3D \cite{chen_nowcast3d_2026} &
    Diffusion &
    Event-curated; filtering unclear &
    Equivalent to training &
    No &
    --- \\

    \addlinespace
    PixelFlowCast \cite{zhu_pixelflowcast_2026} &
    Flow matching &
    Event-curated / unclear &
    Equivalent to training &
    No &
    --- \\

    \addlinespace
    RainDiff \cite{nguyen_raindiff_2025} &
    Diffusion &
    Event-curated / uniform / unclear &
    Equivalent to training &
    Mixed &
    --- \\

    \addlinespace
    HARECast \cite{wen_harecast_2026} &
    Diffusion &
    Event-curated / deterministic filter &
    Equivalent to training &
    No &
    --- \\

    \addlinespace
    EchoCast-3D \cite{wan_echocast3d_2026} &
    Diffusion &
    Uniform (warm season) &
    Equivalent to training; extreme-event subset &
    Mixed &
    Coverage / spread--skill only \\

    \addlinespace
    CRFT \cite{zhou_crft_2026} &
    Flow matching &
    Event-curated / deterministic filtering &
    Equivalent to training &
    No &
    --- \\

    \addlinespace
    NowcastDiff \cite{liu_nowcastdiff_2026} &
    Diffusion &
    Event-curated; details unclear &
    Equivalent to training &
    No &
    --- \\

    \addlinespace
    cGAN-att \cite{piran_cGAN-att_2024} &
    GAN &
    Event-curated; mechanism unclear &
    Equivalent to training &
    No &
    --- \\

    \addlinespace
    RadarDiT \cite{wan_radardit_2025} &
    Diffusion &
    Event-curated; details unclear &
    Equivalent to training &
    No &
    Spread--skill only \\

    \addlinespace
    CastDiffuser \cite{niu_castdiffuser_2026} &
    Diffusion &
    Event-curated &
    Equivalent to training &
    No &
    --- \\

    \addlinespace
    RNDiff \cite{ling_rndiff_2025} &
    Diffusion &
    Deterministic filter &
    Equivalent to training &
    No &
    --- \\

    \addlinespace
    Physics-guided LDM \cite{gao_physics-guided_2026} &
    Diffusion &
    Event-curated &
    Equivalent to training &
    No &
    --- \\

    \addlinespace
    HyDiff \cite{sun_hydiff_2025} &
    Diffusion &
    Deterministic filter &
    Equivalent to training &
    No &
    --- \\

    \addlinespace
    Spatiotemporal Transformer DDPM \cite{zhao_advancing_2024} &
    Diffusion &
    Event-curated &
    Equivalent to training &
    No &
    --- \\

    \addlinespace
    MTLDM \cite{li_mtldm_2025} &
    Diffusion &
    Chronological; construction unclear &
    Equivalent to training &
    Unclear &
    --- \\

    \addlinespace
    Forecastformer \cite{geng_forecastformer_2025} &
    Diffusion &
    Uniform source; additional selection unclear &
    Equivalent to training &
    Unclear &
    --- \\

    \addlinespace
    HybridFlow \cite{ling_hybridflow_2026} &
    Flow matching &
    Event-curated; details unclear &
    Equivalent to training &
    No &
    --- \\

    \addlinespace
    MS-EdgeCast \cite{zhang_ms-edgecast_2026} &
    Diffusion &
    Deterministic filter &
    Equivalent to training &
    No &
    --- \\

    \addlinespace
    DTCA \cite{li_dtca_2024} &
    Diffusion &
    Unclear &
    Unclear &
    Unclear &
    --- \\

    \addlinespace
    SSLDM-ISI \cite{ling_ssldm_2024} &
    Diffusion &
    Chronological; construction unclear &
    Equivalent to training &
    Unclear &
    --- \\

    \addlinespace
    TRDM \cite{ling_trdm_2024} &
    Diffusion &
    Chronological; construction unclear &
    Equivalent to training &
    Unclear &
    --- \\

    \addlinespace
    STLDM \cite{foo_stldm_2025} &
    Diffusion &
    Event-curated / deterministic filtering &
    Equivalent to training &
    No &
    --- \\

\end{longtable}

\endgroup

\subsection{Importance Sampling Configuration and Sensitivity}

This section lists the parameter configuration of the importance sampling in Supplementary Table \ref{tab:importance-sampling-parameters} and studies their sensitivity in Supplementary Fig. \ref{fig:sampling-parameter-sensitivity} and Supplementary Table \ref{tab:sampling-rain-class-fractions}.

\begin{table}[htbp]
    \centering
    \caption{Parameters of the target-dependent importance sampling scheme used to construct the training set. The normalization size $D$ is the number of radar values in one target crop.}
    \label{tab:importance-sampling-parameters}
    \small
    \setlength{\tabcolsep}{7pt}
    \begin{tabular}{llr}
        \toprule
        Parameter & Symbol & Value \\
        \midrule
        Saturation scale & $s$ & 1\,mm\,h$^{-1}$ \\
        Rain contribution coefficient & $m$ & 2 \\
        Minimum acceptance probability & $q_{\min}$ & 0.004 \\
        Time steps & $T_y$ & 3 \\
        Crop height & $H$ & 128 pixels \\
        Crop width & $W$ & 128 pixels \\
        Normalization size & $D=T_y\,H\,W$ & 49,152 \\
        \bottomrule
    \end{tabular}
\end{table}

\begin{figure}[htbp]
    \centering
    \includegraphics[width=\textwidth]{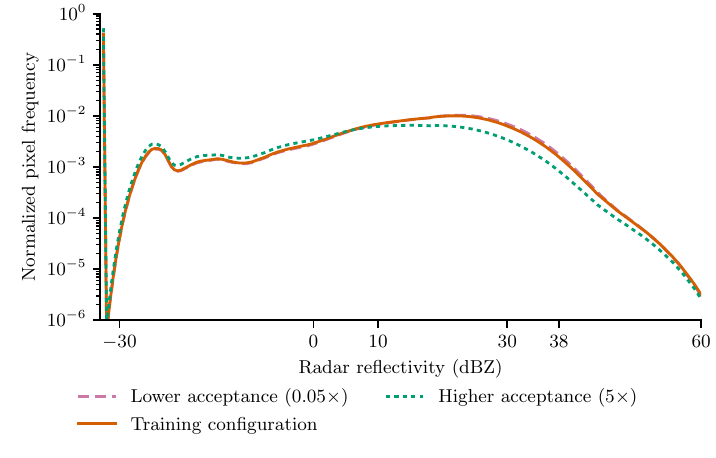}
    \caption{Sensitivity of the pooled training-target distribution to the effective acceptance coefficients. Curves show normalized radar-reflectivity frequencies across all target pixels retained under the training configuration in Supplementary Table~\ref{tab:importance-sampling-parameters} and two sensitivity settings. The lower-acceptance setting scales $m$ and $q_{\min}$ by $0.05$ relative to the training configuration, whereas the higher-acceptance setting scales both by $5$; all other parameters are unchanged. Higher acceptance causes a larger fraction of candidate crops to reach unit acceptance probability, reduces the relative enrichment of rainy crops, and shifts the retained distribution toward no-rain values.}
    \label{fig:sampling-parameter-sensitivity}
\end{figure}

\begin{table}[htbp]
    \centering
    \caption{Radar-reflectivity class fractions in the retained training crops under the training configuration and two sensitivity settings. Values are percentages of all pooled target pixels and sum to 100\% within each row. The lower- and higher-acceptance settings multiply both $m$ and $q_{\min}$ by $0.05$ and $5$, respectively, relative to Supplementary Table~\ref{tab:importance-sampling-parameters}, while holding $s$, $T_y$, $H$, and $W$ fixed. Class boundaries are left-inclusive and right-exclusive except for the open outer bounds.}
    \label{tab:sampling-rain-class-fractions}
    \small
    \setlength{\tabcolsep}{3.8pt}
    \resizebox{\textwidth}{!}{%
    \begin{tabular}{lrrrrr}
        \toprule
        Setting & \shortstack{No rain\\($<10$\,dBZ)}
        & \shortstack{Light\\($10$--$30$\,dBZ)}
        & \shortstack{Moderate\\($30$--$38$\,dBZ)}
        & \shortstack{Strong\\($38$--$48$\,dBZ)}
        & \shortstack{Extreme\\($\geq48$\,dBZ)} \\
        \midrule
        Lower-acceptance sensitivity ($0.05\times$) & 56.251 & 35.519 & 6.852 & 1.281 & 0.098 \\
        Training configuration & 57.788 & 34.767 & 6.188 & 1.160 & 0.097 \\
        Higher-acceptance sensitivity ($5\times$) & 72.465 & 23.538 & 3.270 & 0.655 & 0.071 \\
        \bottomrule
    \end{tabular}%
    }
\end{table}

\subsection{Data Processing Details} \label{sec:data-processing-details}

\begin{table}[htbp]
    \centering
    \caption{Effect of quality control on sample count for both domains. Timestamps refer to the 15-min grid of the May--October seasons. A timestamp is globally invalid if the radar file is missing, if at least one of the SEVIRI channel files is missing so that the scene cannot be gridded, or if a gridded channel contains missing values. Crop-timestamps are timestamps multiplied by the number of lattice crops (91 over Germany, 300 over Italy). A crop-timestamp is masked if its timestamp is globally invalid or if more than 360 of its radar pixels are missing. Candidate sequences are six-frame samples for which all frames are valid and fall within May--October.}
    \label{tab:qc-counts}
    \small
    \setlength{\tabcolsep}{3.8pt}
    \begin{tabular}{lrr}
        \toprule
        Quantity & Germany (2015--2020) & Italy (2025) \\
        \midrule
        Timestamps on the 15-min grid & 105,984 & 17,664 \\
        Globally invalid timestamps & 807 & 30 \\
        \quad missing satellite source files & 690 & 16 \\
        \quad missing satellite values & 35 & 14 \\
        \quad missing radar files & 82 & 0 \\
        Crop-timestamps & 9,644,544 & 5,299,200 \\
        Masked crop-timestamps & 217,121 (2.25\%) & 3,275,540 (61.8\%) \\
        Candidate sequences, training & 6,221,943 & --- \\
        Candidate sequences, validation & 1,568,930 & --- \\
        Candidate sequences, test & 1,543,127 & 1,967,513 \\
        \bottomrule
    \end{tabular}
\end{table}

\begin{table}[htbp]
    \centering
    \caption{Sizes of all datasets used in this study. Importance sampled sets use Eq.~\eqref{eq:acceptanceProb}, while uniform sets are simple random samples without replacement of the same size as the corresponding importance sampled set. The acceptance rate is the size of the importance sampled set relative to the candidate pool. Only the uniform validation set is used for checkpoint selection.}
    \label{tab:dataset-sizes}
    \small
    \setlength{\tabcolsep}{3.8pt}
    \begin{tabular}{lrrrr}
        \toprule
        Split (season) & Candidates & Importance sampled & Uniform & Acceptance rate \\
        \midrule
        Training (2015--2018) & 6,221,943 & 425,713 & --- & 6.84\% \\
        Validation (2019) & 1,568,930 & 125,765 & 125,765 & 8.02\% \\
        Test Germany (2020) & 1,543,127 & 122,644 & 122,644 & 7.95\% \\
        Test Italy (2025) & 1,967,513 & 95,111 & 95,111 & 4.83\% \\
        \bottomrule
    \end{tabular}
\end{table}

Supplementary Table~\ref{tab:qc-counts} lists the counts removed by each quality-control rule and Supplementary Table~\ref{tab:dataset-sizes} the sizes of all datasets. The German radar fields are stored standardized with preliminary statistics and are re-standardized on loading with the values in Supplementary Table~\ref{tab:model-hyperparameters}. Standardized fields are stored in 16-bit floating point and read as 32-bit numbers. The Italian ground truth is stored as $\log(1+R)$ standardized with mean 0.3147 and standard deviation 0.5722 and is transformed back to mm\,h$^{-1}$ and clipped at zero before verification.

The mean inverse acceptance weight $1/a(y)$ is 14.60 on the German training set, which normalizes the IPW loss, 12.54 on the German importance sampled test set, and 20.70 on the Italian one. The corresponding weight sums of 1,538,438 and 1,969,183 estimate the candidate-pool sizes of 1,543,127 and 1,967,513 in Supplementary Table~\ref{tab:qc-counts}.

Data processing used wradlib 2.4.0 \cite{heistermann_technical_2013}, satpy 0.55.0 \cite{raspaud_pytroll_2018}, pyresample 1.33.0, and zarr 2.16.1.

\subsection{Hyperparameters and Training Details} \label{sec:hyperparameters}

This section lists the hyperparameters in Supplementary Table \ref{tab:model-hyperparameters}, presents the value chain for one training step of the model in Supplementary Fig. \ref{fig:valueChain}, while additional details for architecture are displayed in Supplementary Figs. \ref{fig:unet_architecture} and \ref{fig:diffusion_conditioning}.

\begin{figure}[htbp]
    \centering
    \includegraphics[
        width=\textwidth,
        height=0.9\textheight,
        keepaspectratio
    ]{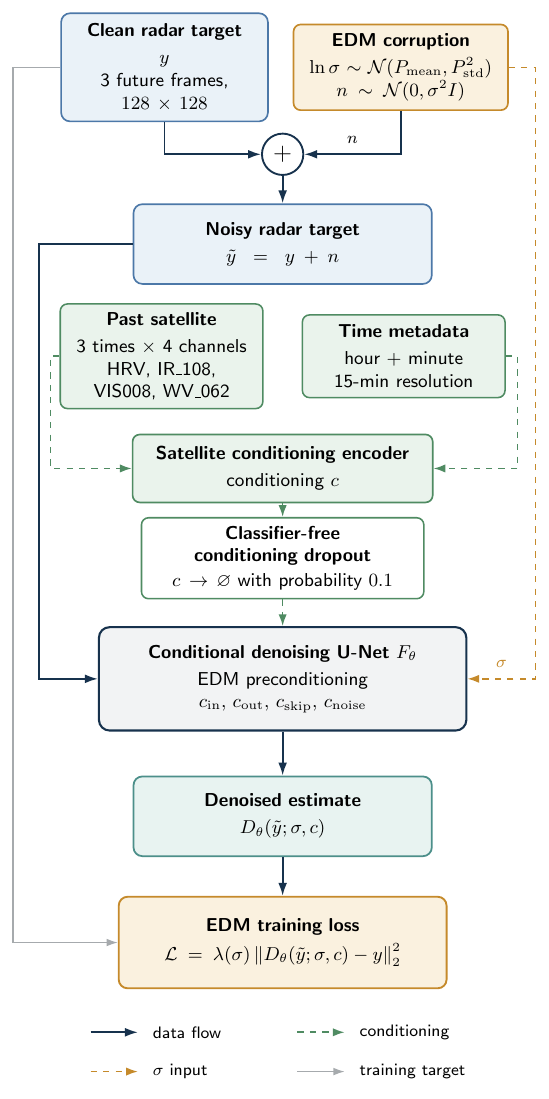}
    \caption{Schematic of one EDM training step in the notation of Section~\ref{sec:model}. }
    \label{fig:valueChain}
\end{figure}

\begin{figure}[htbp]
    \centering
    \includegraphics[
        width=\textwidth,
        height=\textheight,
        keepaspectratio
    ]{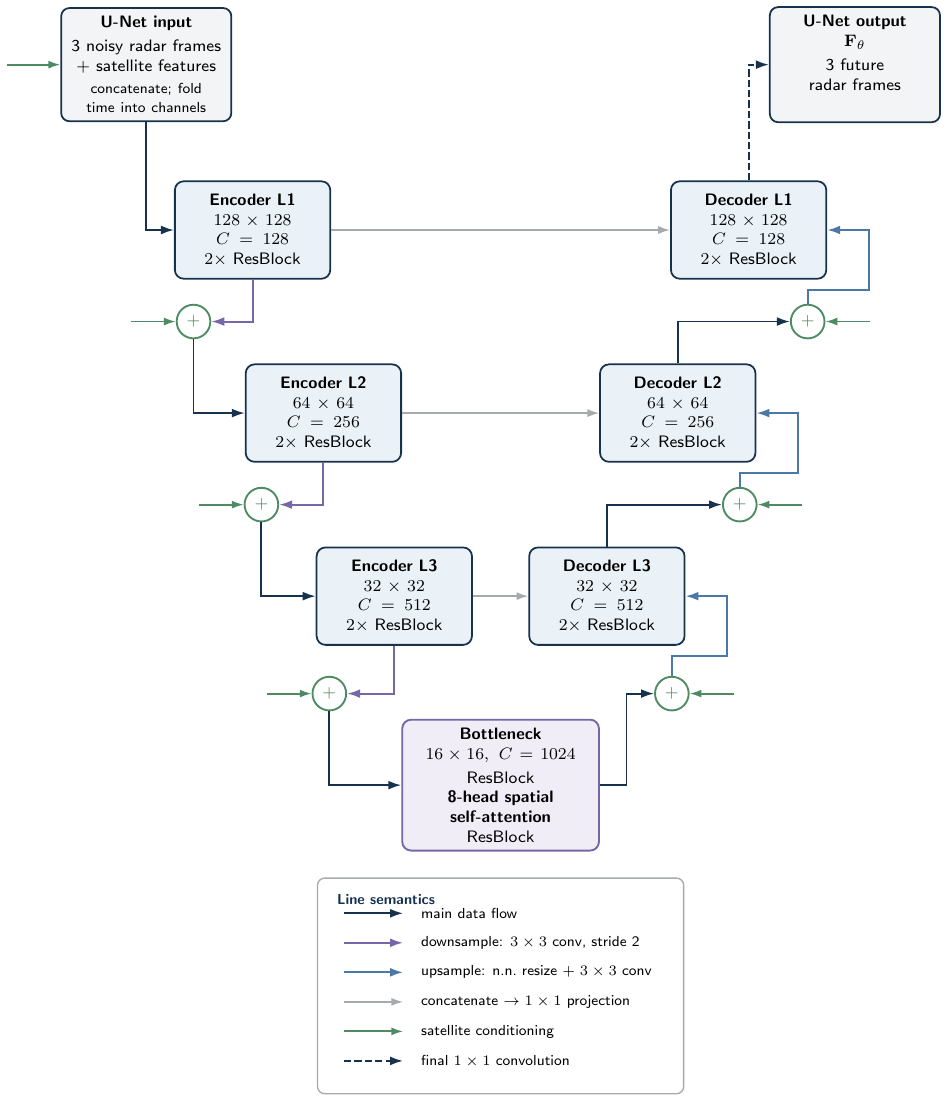}
    \caption{
        High-level architecture of the conditional 2D U-Net $F_\theta$. The three noisy radar frames are concatenated with the full-resolution satellite features and the temporal dimension is folded into the channel dimension. Encoder and decoder operate on four resolution levels from $128\times128$ to $16\times16$ pixels with 128 to 1024 channels and two residual blocks per level; the bottleneck additionally contains an 8-head spatial self-attention block. Downsampling uses strided $3\times3$ convolutions and upsampling nearest-neighbour resizing followed by a $3\times3$ convolution. Skip connections are concatenated and reduced by a $1\times1$ projection. Multi-scale satellite conditioning (green arrows) is added after each downsampling and before each upsampling. A final $1\times1$ convolution yields the three output frames.
    }
    \label{fig:unet_architecture}
\end{figure}

\begin{figure}[htbp]
    \centering

    \begin{subfigure}[o]{0.50\textwidth}
        \centering
        \includegraphics[
            width=\linewidth,
            trim=0 0 0 0,
            clip
        ]{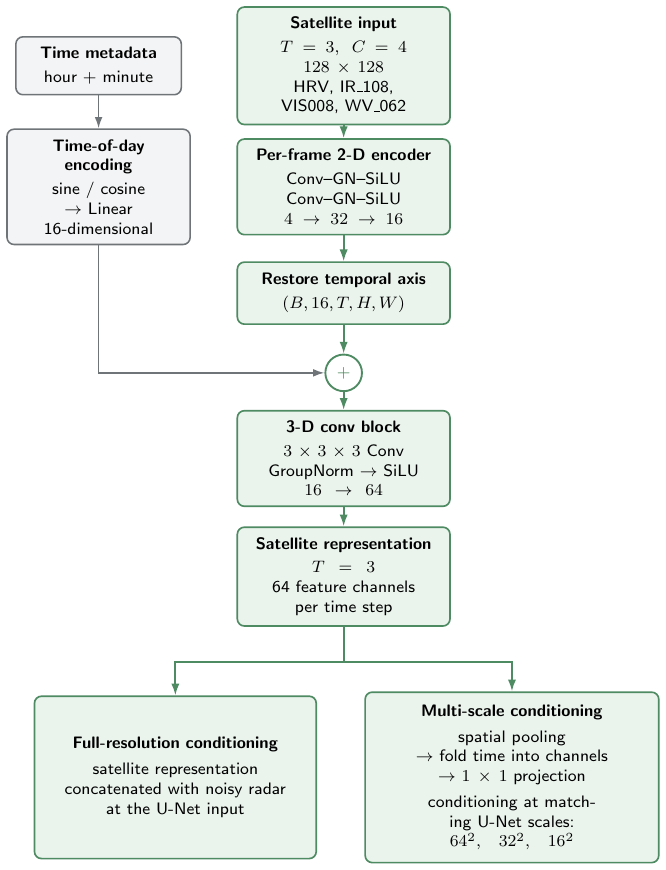}
        \caption{Satellite conditioning encoder.}
        \label{fig:diffusion_arch_satellite}
    \end{subfigure}
    \hfill
    \begin{subfigure}[htbp]{0.43\textwidth}
        \centering
        \includegraphics[
            width=\linewidth,
            trim=0 0 0 0,
            clip
        ]{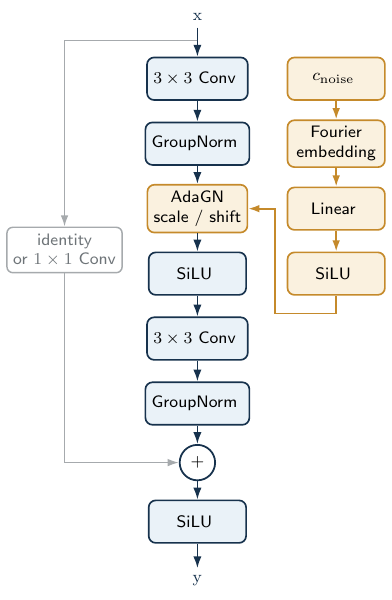}
        \caption{Diffusion embedding and conditioned residual block.}
        \label{fig:diffusion_arch_resblock}
    \end{subfigure}

    \caption{
        Conditioning components of the conditional diffusion model.
        \textbf{(a)} Satellite conditioning encoder. Each of the three input frames with four channels passes through two convolution blocks (convolution, group normalization, SiLU) with 32 and 16 channels. A 16-dimensional sine--cosine encoding of the time of day of each frame, followed by a linear layer, is added to the per-frame features. After restoring the temporal axis, a $3\times3\times3$ convolution block fuses the frames into 64 feature channels per time step. This representation is concatenated with the noisy radar at the U-Net input (full-resolution conditioning) and, after spatial pooling, folding time into channels, and a $1\times1$ projection, added at the $64^2$, $32^2$, and $16^2$ levels (multi-scale conditioning).
        \textbf{(b)} Conditioned residual block. The noise level $c_{\mathrm{noise}}$ is mapped by a Fourier embedding, a linear layer, and SiLU to channel-wise scale and shift parameters that modulate the group-normalized activations (adaptive GroupNorm, AdaGN). The block consists of two $3\times3$ convolutions with group normalization and SiLU and a residual connection that is the identity or a $1\times1$ convolution when the channel count changes.
    }
    \label{fig:diffusion_conditioning}
\end{figure}

\begingroup
\footnotesize
\setlength{\LTleft}{0pt}
\setlength{\LTright}{0pt}
\setlength{\tabcolsep}{5pt}
\renewcommand{\arraystretch}{1.12}
\begin{longtable}{@{}>{\raggedright\arraybackslash}p{0.31\textwidth}>{\raggedright\arraybackslash}p{0.65\textwidth}@{}}
    \caption{Hyperparameters used for model training and ensemble inference. The
    baseline and IPW models differ only in the per-sample loss multiplier. The model was implemented in PyTorch 2.5.1 \cite{paszke_pytorch_2019} with PyTorch Lightning. Values
    labelled as PyTorch defaults are stated explicitly. The epoch is zero indexed. Training and inference are seeded, but bitwise reproducibility is not guaranteed because deterministic kernels were not enforced and non-deterministic GPU algorithms were not disabled. All computation is in 32-bit floating point.}
    \label{tab:model-hyperparameters}\\
    \toprule
    Hyperparameter & Value \\
    \midrule
    \endfirsthead
    \multicolumn{2}{@{}l}{\footnotesize\textit{\tablename~\thetable\ continued}}\\
    \addlinespace
    \toprule
    Hyperparameter & Value \\
    \midrule
    \endhead
    \bottomrule
    \endfoot
    \multicolumn{2}{@{}l}{\textbf{Data and preprocessing}}\\*
    \addlinespace[2pt]
    Satellite predictors & HRV, IR\_108, VIS008, WV\_062 \\
    Input/target sequence & 3 input frames ($-30,-15,0$ min); 3 target frames ($0,15,30$ min) \\
    Temporal resolution & 15 min \\
    Crop size and stride & $128\times128$ pixels; 64-pixel stride (50\% overlap) \\
    Satellite standardization $(\mu,\sigma)$ & HRV $(11.3462,14.2666)$; IR\_108 $(270.7699,16.8125)$; VIS008 $(12.9961,15.3883)$; WV\_062 $(232.0065,4.3216)$ \\
    Radar standardization $(\mu,\sigma)$ & $(-3.9248,25.7472)$ dBZ \\
    Sampling used for training/validation & Importance sampling / uniform validation \\
    Training/validation samples & 425,713 / 125,765 crops (3,326 optimization steps per epoch) \\
    \addlinespace[5pt]
    \multicolumn{2}{@{}l}{\textbf{Network}}\\*
    \addlinespace[2pt]
    Architecture & Conditional U-Net; 82,297,571 trainable parameters \\
    Input/output channels & 4 satellite and 1 noisy-radar / 1 radar \\
    Satellite feature width & 64 (2-D widths 32, 16; subsequent 3-D width 64) \\
    U-Net channel widths & 128, 256, 512, 1024 \\
    Residual blocks & 2 per encoder and decoder level; 2 in bottleneck \\
    Bottleneck attention & 8-head spatial self-attention at $16\times16$ pixels \\
    Normalization / activation & Group normalization (32 groups; 8 in satellite encoder) / SiLU \\
    Noise embedding & 128-dimensional random Fourier features; frequency scale 16 \\
    Time-of-day encoding & 16-dimensional sinusoidal encoding of the fraction of day of each input frame; linear layer; added to the per-frame satellite features before the 3-D convolution \\
    Multi-scale conditioning & Satellite features average pooled by factors 2, 4, 8 and projected to 256, 512, 1024 channels; added after each downsampling and before each upsampling \\
    Unconditional branch & Satellite encoder and all injections skipped; dedicated first-level stem (513,920 parameters); all deeper levels shared \\
    Upsampling & Nearest-neighbour resize followed by $3\times3$ convolution \\
    \addlinespace[5pt]
    \multicolumn{2}{@{}l}{\textbf{EDM objective}}\\*
    \addlinespace[2pt]
    Data-noise scale & $\sigma_{\mathrm{data}}=1.0$ \\
    Training-noise distribution & $\ln\sigma\sim\mathcal{N}(P_{\mathrm{mean}},P_{\mathrm{std}}^2)$ \\
    Noise-distribution parameters & $P_{\mathrm{mean}}=-0.5$, $P_{\mathrm{std}}=1.2$ \\
    Loss & EDM-preconditioned, noise-level-weighted mean squared error \\
    Baseline / IPW multiplier & $1$ / $[1/a(y)]/14.6028$ \\
    \addlinespace[5pt]
    \multicolumn{2}{@{}l}{\textbf{Optimization}}\\*
    \addlinespace[2pt]
    Optimizer & AdamW \\
    AdamW $\beta_1,\beta_2$; $\epsilon$; weight decay & $(0.9,0.999)$; $10^{-8}$; 0.01 (PyTorch defaults) \\
    Batch size / gradient accumulation & 128 / 1 \\
    Initial learning rate & $1.5\times10^{-4}$ \\
    Learning-rate schedule & Linear warm-up for 5 epochs, then cosine annealing for 95 epochs \\
    Warm-up start / cosine minimum & $10^{-6}$ / $5\times10^{-6}$ \\
    Epoch cap / selected checkpoint & 100 / epoch 81 (baseline and IPW) \\
    Early stopping & Uniform-validation EDM loss; patience 10; minimum change 0 \\
    Gradient clipping & Global norm 1.0 \\
    Exponential moving average & Decay 0.999 from step 0; EMA weights used for validation and inference \\
    Classifier-free conditioning dropout & 0.10 \\
    Spatial augmentation & Random $90^\circ$ rotation and independent horizontal/vertical flips ($p=0.5$ each) \\
    Training seed / numerical precision / device & 31 / 32-bit / single NVIDIA HGX A100 80GB 500W GPU \\
    \addlinespace[5pt]
    \multicolumn{2}{@{}l}{\textbf{Ensemble inference}}\\*
    \addlinespace[2pt]
    Sampler & DPM-Solver++(2M) with the EDM (Karras) noise schedule \\
    Sampling steps & 50 \\
    Sampling-noise range and schedule exponent & $\sigma_{\max}=80$, $\sigma_{\min}=0.002$, $\rho=7$ \\
    Initial latent & Independent $\mathcal{N}(0,\sigma_{\max}^2 I)$ noise for each member \\
    Ensemble size & 20 \\
    Classifier-free guidance scales & 1, 2, and 4 (scale 1 for the baseline/IPW comparison) \\
    Network evaluations per member & 50 at guidance scale 1; 100 at scales 2 and 4 \\
    Inference seed / batch size & 8932 / 128 \\
\end{longtable}
\endgroup

\subsection{Verification Details} \label{sec:verification-details}

The rank histogram, CRPS, and power spectra are computed with pysteps 1.12.0. \cite{pulkkinen_pysteps_2019}; the FSS re-implements the pysteps smoothing so that numerator and denominator of Eq.~\eqref{eq:fss} can be accumulated across samples and blocks. The random tie-breaking of the rank histogram uses an unseeded random number generator, so repeated evaluations differ by the tie-breaking noise. This affects only how the tied no-rain mass is distributed across neighboring ranks and does not alter any conclusion, but the histograms are not bitwise reproducible. The block bootstrap uses a fixed seed; the 27 seven-day blocks span 26 April to 25 October 2020 over Germany and 27 April to 26 October 2025 over Italy, i.e.\ block boundaries are fixed by the calendar rather than by the data. No confidence intervals are computed for the power spectra.

\subsection{Additional Evaluation}

\begin{figure}[htbp]
    \centering
    \includegraphics[width=\textwidth]{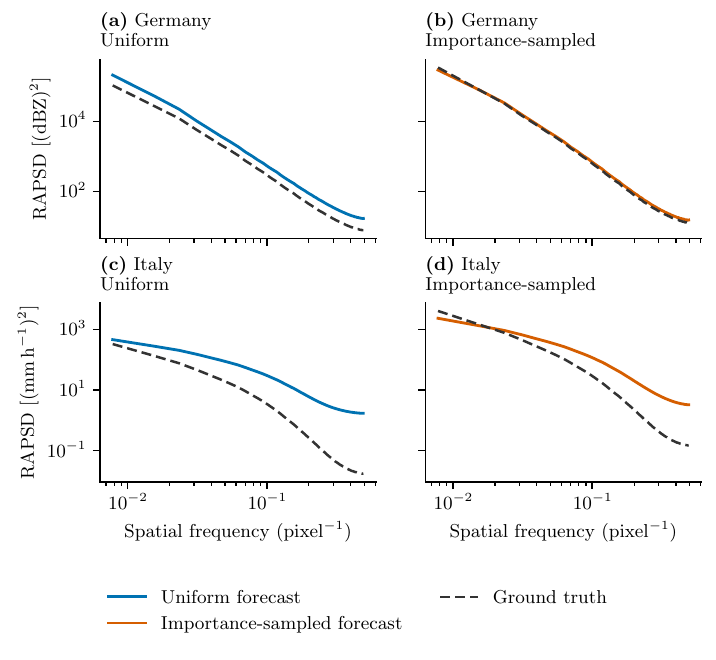}
    \caption{Radially averaged power spectral density (RAPSD) of the model forecasts and corresponding ground-truth fields over Germany (top row) and Italy (bottom row), evaluated on the uniform (left column) and importance sampled (right column) test sets. Curves aggregate all forecast lead times. Forecast spectra use the color of the respective sampling protocol, while ground truth is shown by the charcoal dashed curves. Both axes are logarithmic, and spatial frequency is given in pixel$^{-1}$. The absolute spectral densities should not be compared across domains because the German fields contain radar reflectivity in dBZ, whereas the Italian fields contain precipitation rates in mm\,h$^{-1}$. Notably, the model reproduces the RAPSD almost perfectly when evaluated on the importance sampled German set. However, the model lies consistently above the ground truth on the uniform set. Furthermore, the RAPSD does not generalize well to the Italian domain on either set.}
    \label{fig:rapsd-test-set-comparison}
\end{figure}

\begin{figure}[htbp]
    \centering
    \includegraphics[width=\textwidth]{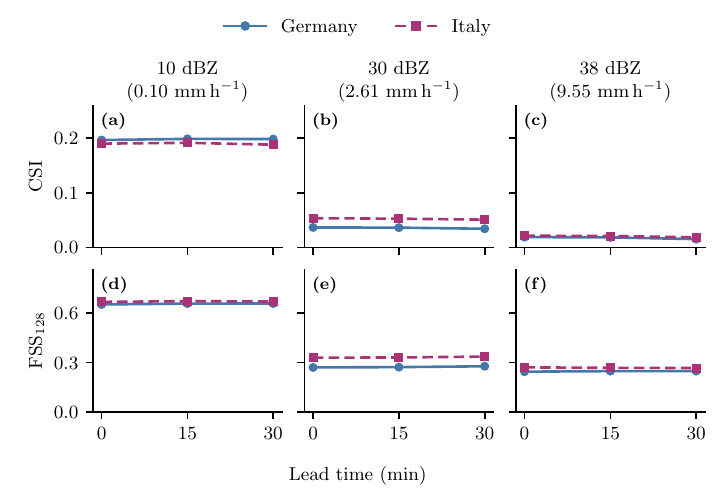}
    \caption{Critical success index (CSI; top row) and fractions skill score at a neighborhood scale of 128 pixels ($\mathrm{FSS}_{128}$; bottom row) as functions of lead time on the uniform German and Italian test sets. Columns correspond to threshold-matched intensity levels of 10, 30, and 38\,dBZ over Germany and 0.10, 2.61, and 9.55\,mm\,h$^{-1}$ over Italy, respectively. Scores are member averaged; solid blue curves with circles denote Germany and dashed magenta curves with squares denote Italy. Across the available 0--30-minute range, both scores are nearly invariant with lead time, apart from a modest decrease in CSI at the highest threshold. This weak temporal degradation is consistent with the satellite-to-radar conversion being the dominant difficulty over these short lead times, rather than forecast extrapolation alone.}
    \label{fig:skill-vs-lead-time}
\end{figure}

This section presents additional evaluation results. Supplementary Figure \ref{fig:rapsd-test-set-comparison} displays the radially averaged power spectral density (RAPSD) on both domains and test set sample protocols, while Supplementary Fig. \ref{fig:skill-vs-lead-time} shows the lead time dependence of selected skill scores. Complementary verification scores for Table~\ref{tab:test-set-score-comparison} are in Supplementary Table \ref{tab:complement-test-set-score-comparison}, while complementary verification scores for the classifier-free guidance comparison are in Supplementary Table \ref{tab:complement-guidance-score-comparison}. 

\begin{table}[htbp]
    \centering
    \caption{Complementary verification scores for Table~\ref{tab:test-set-score-comparison}. Definitions, aggregation, threshold matching, evaluation protocols, and confidence intervals are as in Table~\ref{tab:test-set-score-comparison}. FSS is shown at the two neighborhood scales omitted from the main table. CRPS is reported in dBZ for Germany and mm\,h$^{-1}$ for Italy and is not comparable across domains.}
    \label{tab:complement-test-set-score-comparison}
    \small
    \setlength{\tabcolsep}{3.5pt}
    \begin{tabular}{llccc}
        \toprule
        Metric & Threshold & \shortstack{Uniform\\test set}
        & \shortstack{Importance sampled\\test set}
        & \shortstack{IPW estimate from\\importance set} \\
        \midrule
        \multicolumn{5}{l}{\textit{Germany}} \\
        \multirow{3}{*}{POD}
        & 10\,dBZ & 0.420 [0.411, 0.431] & 0.450 [0.439, 0.460] & 0.421 [0.411, 0.431] \\
        & 30\,dBZ & 0.060 [0.050, 0.072] & 0.047 [0.038, 0.058] & 0.059 [0.048, 0.072] \\
        & 38\,dBZ & 0.029 [0.022, 0.036] & 0.019 [0.013, 0.024] & 0.028 [0.021, 0.035] \\
        \addlinespace[1pt]
        \multirow{3}{*}{FAR}
        & 10\,dBZ & 0.727 [0.700, 0.756] & 0.347 [0.322, 0.375] & 0.727 [0.701, 0.755] \\
        & 30\,dBZ & 0.919 [0.900, 0.939] & 0.809 [0.785, 0.840] & 0.919 [0.901, 0.938] \\
        & 38\,dBZ & 0.958 [0.946, 0.972] & 0.919 [0.906, 0.938] & 0.958 [0.947, 0.971] \\
        \addlinespace[1pt]
        \multirow{3}{*}{$\mathrm{FSS}_{8}$}
        & 10\,dBZ & 0.422 [0.398, 0.444] & 0.600 [0.589, 0.611] & 0.423 [0.399, 0.443] \\
        & 30\,dBZ & 0.121 [0.100, 0.143] & 0.119 [0.099, 0.142] & 0.121 [0.099, 0.143] \\
        & 38\,dBZ & 0.077 [0.058, 0.093] & 0.063 [0.046, 0.078] & 0.076 [0.057, 0.091] \\
        \addlinespace[1pt]
        \multirow{3}{*}{$\mathrm{FSS}_{32}$}
        & 10\,dBZ & 0.526 [0.500, 0.550] & 0.686 [0.676, 0.696] & 0.527 [0.502, 0.549] \\
        & 30\,dBZ & 0.188 [0.155, 0.223] & 0.169 [0.138, 0.206] & 0.186 [0.151, 0.223] \\
        & 38\,dBZ & 0.151 [0.112, 0.184] & 0.114 [0.082, 0.148] & 0.146 [0.107, 0.181] \\
        \addlinespace[1pt]
        CRPS & --- & 12.21 [12.04, 12.37] & 11.97 [11.78, 12.14] & 12.21 [12.03, 12.40] \\
        \midrule
        \multicolumn{5}{l}{\textit{Italy}} \\
        \multirow{3}{*}{POD}
        & 10\,dBZ & 0.392 [0.382, 0.400] & 0.439 [0.432, 0.446] & 0.392 [0.383, 0.400] \\
        & 30\,dBZ & 0.084 [0.075, 0.094] & 0.079 [0.071, 0.088] & 0.086 [0.077, 0.094] \\
        & 38\,dBZ & 0.037 [0.031, 0.045] & 0.029 [0.024, 0.034] & 0.037 [0.031, 0.042] \\
        \addlinespace[1pt]
        \multirow{3}{*}{FAR}
        & 10\,dBZ & 0.730 [0.710, 0.755] & 0.349 [0.334, 0.363] & 0.729 [0.707, 0.753] \\
        & 30\,dBZ & 0.879 [0.866, 0.891] & 0.729 [0.711, 0.747] & 0.875 [0.861, 0.888] \\
        & 38\,dBZ & 0.958 [0.949, 0.966] & 0.920 [0.910, 0.930] & 0.957 [0.949, 0.964] \\
        \addlinespace[1pt]
        \multirow{3}{*}{$\mathrm{FSS}_{8}$}
        & 10\,dBZ & 0.415 [0.394, 0.432] & 0.592 [0.585, 0.599] & 0.416 [0.396, 0.434] \\
        & 30\,dBZ & 0.157 [0.142, 0.171] & 0.172 [0.156, 0.187] & 0.160 [0.145, 0.173] \\
        & 38\,dBZ & 0.076 [0.065, 0.087] & 0.074 [0.064, 0.084] & 0.077 [0.066, 0.087] \\
        \addlinespace[1pt]
        \multirow{3}{*}{$\mathrm{FSS}_{32}$}
        & 10\,dBZ & 0.536 [0.512, 0.556] & 0.695 [0.687, 0.703] & 0.537 [0.512, 0.558] \\
        & 30\,dBZ & 0.242 [0.217, 0.266] & 0.237 [0.213, 0.260] & 0.245 [0.222, 0.268] \\
        & 38\,dBZ & 0.159 [0.133, 0.186] & 0.144 [0.120, 0.169] & 0.162 [0.137, 0.185] \\
        \addlinespace[1pt]
        CRPS & --- & 0.079 [0.064, 0.094] & 0.872 [0.791, 0.953] & 0.080 [0.065, 0.095] \\
        \bottomrule
    \end{tabular}
\end{table}

\begin{table}[htbp]
    \centering
    \caption{Complementary verification scores for the classifier-free guidance comparison in Table~\ref{tab:guidance-score-comparison}. The evaluation set, aggregation, and confidence intervals are as in Table~\ref{tab:guidance-score-comparison}. FSS is shown at the two neighborhood scales omitted from the main table; CRPS is reported in dBZ.}
    \label{tab:complement-guidance-score-comparison}
    \small
    \setlength{\tabcolsep}{3.5pt}
    \begin{tabular}{llccc}
        \toprule
        Metric & Threshold & Guidance scale 1 & Guidance scale 2 & Guidance scale 4 \\
        \midrule
        \multirow{3}{*}{POD}
        & 10\,dBZ & 0.420 [0.411, 0.431] & 0.523 [0.505, 0.541] & 0.609 [0.589, 0.631] \\
        & 30\,dBZ & 0.060 [0.050, 0.072] & 0.161 [0.135, 0.190] & 0.301 [0.264, 0.338] \\
        & 38\,dBZ & 0.029 [0.022, 0.036] & 0.107 [0.087, 0.125] & 0.226 [0.196, 0.252] \\
        \addlinespace
        \multirow{3}{*}{FAR}
        & 10\,dBZ & 0.727 [0.700, 0.756] & 0.692 [0.664, 0.720] & 0.704 [0.677, 0.733] \\
        & 30\,dBZ & 0.919 [0.900, 0.939] & 0.904 [0.885, 0.925] & 0.920 [0.904, 0.937] \\
        & 38\,dBZ & 0.958 [0.946, 0.972] & 0.950 [0.937, 0.964] & 0.965 [0.955, 0.975] \\
        \addlinespace
        \multirow{3}{*}{$\mathrm{FSS}_{8}$}
        & 10\,dBZ & 0.422 [0.398, 0.444] & 0.487 [0.461, 0.510] & 0.498 [0.470, 0.524] \\
        & 30\,dBZ & 0.121 [0.100, 0.143] & 0.204 [0.168, 0.236] & 0.202 [0.168, 0.234] \\
        & 38\,dBZ & 0.077 [0.058, 0.093] & 0.146 [0.113, 0.171] & 0.115 [0.085, 0.141] \\
        \addlinespace
        \multirow{3}{*}{$\mathrm{FSS}_{32}$}
        & 10\,dBZ & 0.526 [0.500, 0.550] & 0.595 [0.568, 0.621] & 0.605 [0.575, 0.633] \\
        & 30\,dBZ & 0.188 [0.155, 0.223] & 0.304 [0.252, 0.353] & 0.299 [0.248, 0.344] \\
        & 38\,dBZ & 0.151 [0.112, 0.184] & 0.268 [0.207, 0.313] & 0.194 [0.141, 0.237] \\
        \addlinespace
        CRPS & --- & 12.21 [12.04, 12.37] & 8.46 [8.15, 8.78] & 7.09 [6.46, 7.73] \\
        \bottomrule
    \end{tabular}
\end{table}

\subsection{IPW Training and Post-Processing} \label{sec:IPWTrainingAndPostporcessing}

Section~\ref{sec:forecaster_dilemma} and Table~\ref{tab:test-set-score-comparison} apply IPW at the verification stage, where every test sample is weighted by the inverse of its acceptance probability. Because Eq.~\eqref{eq:conditional-tilt} is invertible, the same correction can in principle be applied at two other stages of the pipeline: to the loss function, so that the model learns the deployment conditional in the first place, or to the samples of the trained model in a post-processing step. Both variants are the identical Horvitz-Thompson reweighting \cite{horvitz_generalization_1952} applied to a different object, namely the loss instead of the score, or the predictive density instead of the test set. Both require that $a(y)$ is known and strictly positive for every candidate sample. This section reports an experiment for the training stage and outlines the post-processing variant.

Weighting the per-sample loss by $w(y)=1/a(y)$ turns the empirical training objective over the importance sampled set into an unbiased estimate of the objective over the deployment distribution, in the same way as importance weights correct a Monte Carlo estimate \cite{kahn_use_1955} or a covariate-shifted training set \cite{shimodaira_improving_2000}. For the EDM loss of Eq.~\eqref{eq:edm-loss} this means that the minimizer approximates the deployment score $\nabla_y \log p_{\mathrm{clim}}(y\,|\,x)$ instead of the shifted score of Eq.~\eqref{eq:score-tilt}. We divide the weights by their mean over the training set (14.60) so that the average loss magnitude, and thereby the effective learning rate, is unchanged, and otherwise train exactly as the baseline (Supplementary Table~\ref{tab:model-hyperparameters}). Both models select epoch 81 by the uniform validation loss. To our knowledge, the only previous account of training-time importance weights in generative nowcasting is a single sentence in the supplement of DGMR, which reports no significant advantage in a GAN setting without further analysis \cite{ravuri_skilful_2021}.

The weights range from 1 for crops with $a(y)=1$, which make up 10.9\% of the training set, to 238 for the driest crops, and the crops with $a(y)<0.01$ carry 57\% of the total weight. The effective sample size $(\sum_i w_i)^2/\sum_i w_i^2$ of the weighted training set is 50,523, i.e.\ 11.9\% of its 425,713 crops. IPW training therefore trades the wet bias of the objective for a much noisier gradient estimate in which rainy crops contribute little.

\begin{figure}[htbp]
    \centering
    \includegraphics[width=\textwidth]{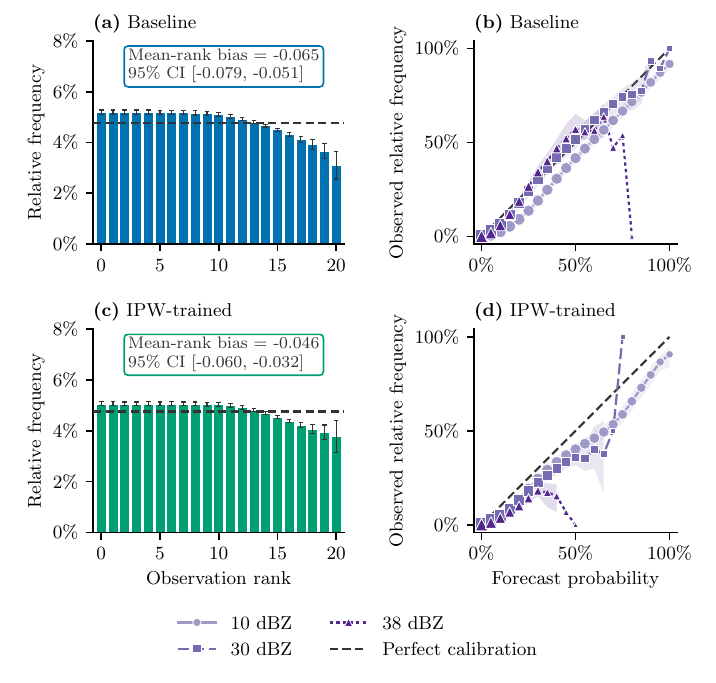}
    \caption{Calibration of the baseline and inverse-probability-weighted (IPW) trained models on the German uniform test set at guidance scale~1. Both models are trained on the same importance sampled training set; the IPW-trained model weights its loss by the inverse sampling acceptance probability. Each row shows one model. The left column contains rank histograms aggregated over all lead times; error bars and the reported signed normalized mean-rank-bias intervals are 95\% confidence intervals from a 10,000-replicate block bootstrap with seven-day blocks. The dashed horizontal line denotes the relative frequency expected under perfect calibration. The right column contains reliability diagrams for exceedance of 10, 30, and 38\,dBZ. Shading denotes the corresponding 95\% block-bootstrap confidence intervals; intervals are omitted where fewer than ten seven-day blocks contribute to a probability bin. Marker area follows the logarithm of the refinement-bin count using one common scale across panels and models, and the diagonal denotes perfect reliability. IPW training moves the rank histogram and mean-rank bias toward perfect calibration, although residual overprediction remains.}
    \label{fig:ipw-training-calibration}
\end{figure}

Supplementary Table~\ref{tab:ipw-training-score-comparison} and Supplementary Fig.~\ref{fig:ipw-training-calibration} compare both models on the German uniform test set at guidance scale 1. Both CRPS measures improve under IPW training. CRPS decreases from 12.21 [12.04, 12.37] to 11.24 [11.05, 11.43]\,dBZ and pooled CRPS from 12.26 [12.00, 12.53] to 11.23 [11.01, 11.44]\,dBZ. The rank histogram flattens and the signed mean-rank bias moves from $-0.065$ [$-0.079$, $-0.051$] to $-0.046$ [$-0.060$, $-0.032$], a paired difference of $+0.019$ [$+0.016$, $+0.022$], while the frequency bias at 10\,dBZ drops from 1.54 to 1.42. The light rain over-prediction of the baseline is thus reduced but not removed. At the same time, categorical skill deteriorates at every threshold and most strongly for strong rain: CSI at 38\,dBZ halves from 0.017 to 0.007, $\mathrm{FSS}_{128}$ falls from 0.247 to 0.145, POD at 30 and 38\,dBZ halves, and FAR rises. Reweighting shrinks the effective sample size of exactly those rare events that the sampling scheme was designed to enrich, so that the model effectively sees them about as rarely as under climatological sampling. 

IPW training is therefore a choice rather than a free correction. If the goal is a calibrated probabilistic product on the deployment distribution, weighting the loss, or training on a climatologically sampled set, is the principled option. If the goal is maximal skill for rare events under a fixed compute budget, the shift may be retained deliberately. Then, however, it must be stated, and the verification must nevertheless be performed on deployment consistent data or with the weighted scores of Section~\ref{sec:verification}. 

The third option leaves the training untouched and corrects the predictive distribution of the trained model. Inspired by \cite{vahid_yousefnia_machine-learning_2024}, inverting Eq.~\eqref{eq:conditional-tilt} and using that the deployment conditional integrates to one gives
\begin{equation}
    p_{\mathrm{clim}}(y\,|\,x) = \frac{p_{\mathrm{train}}(y\,|\,x)/a(y)}{\int p_{\mathrm{train}}(y'\,|\,x)/a(y')\,\mathrm{d}y'},
    \label{eq:ipw-postprocessing}
\end{equation}
so that the unknown normalizer $Z(x)$ is eliminated. For an ensemble of $N_e$ members $y^{(k)}$ drawn from the trained model, any expectation under the deployment conditional can be estimated with the self-normalized member weights
\begin{equation}
    \mathbb{E}_{\mathrm{clim}}\bigl[\varphi(Y)\,|\,x\bigr] \approx \sum_{k=1}^{N_e} \omega^{(k)}\,\varphi\bigl(y^{(k)}\bigr),\qquad
    \omega^{(k)} = \frac{1/a\bigl(y^{(k)}\bigr)}{\sum_{j=1}^{N_e} 1/a\bigl(y^{(j)}\bigr)},
    \label{eq:ipw-member-weights}
\end{equation}
where $\varphi$ may be a threshold exceedance, a pixel value, or any other functional of the forecast. Because $a$ is a known function of the target alone, the weights can be evaluated directly on the generated members without reference to observations. Supplementary Equation~\eqref{eq:ipw-member-weights} is the ensemble analogue of the H\'ajek estimators used for verification, applied within one forecast rather than across the test set.

We have not applied this correction to our model for two reasons, which we consider limitations of the approach rather than of this study. First, Supplementary Equation~\eqref{eq:ipw-postprocessing} is exact only for an ideal model that has learned $p_{\mathrm{train}}(y\,|\,x)$ perfectly. A budget-constrained model deviates from the predicted tilt in a way the theory does not specify (Section~\ref{sec:learningObjectiveTilt}), so the corrected ensemble must still be verified and, if necessary, tuned. This requires a deployment consistent validation set and returns the problem to the verification stage. Second, the member weights are self-normalized within one forecast of only $N_e=20$ members, whereas $1/a(y)$ spans more than two orders of magnitude. Whenever an ensemble contains both dry and rainy members, the driest members receive nearly all of the weight and the effective number of members collapses, so that exceedance probabilities become coarsely quantized and noisy. A reliable estimate of Supplementary Equation~\eqref{eq:ipw-postprocessing} therefore requires far larger ensembles than are common in nowcasting, and probably larger than are affordable for diffusion models at operational latency. Post-processing IPW is thus most attractive for low-dimensional targets or for models whose sampling is cheap. A systematic evaluation is left for future work.

\begin{table}[htbp]
    \centering
    \caption{Verification of the baseline and inverse-probability-weighted (IPW) trained models on the German uniform test set at guidance scale~1. Values in brackets are 95\% confidence intervals from a 10,000-replicate block bootstrap with seven-day blocks. Categorical scores and fractions skill score (FSS) are member averaged and aggregated over all lead times; Pooled CRPS and CRPS are reported in dBZ. The baseline is repeated from Table~\ref{tab:test-set-score-comparison} to make the model comparison self-contained.}
    \label{tab:ipw-training-score-comparison}
    \small
    \setlength{\tabcolsep}{3.5pt}
    \begin{tabular}{llcc}
        \toprule
        Metric & Threshold & Baseline & IPW-trained \\
        \midrule
        \multirow{3}{*}{Bias}
        & 10\,dBZ & 1.541 [1.411, 1.705] & 1.416 [1.295, 1.570] \\
        & 30\,dBZ & 0.747 [0.643, 0.894] & 0.670 [0.578, 0.804] \\
        & 38\,dBZ & 0.687 [0.580, 0.856] & 0.673 [0.565, 0.856] \\
        \addlinespace
         \multirow{3}{*}{POD}
        & 10\,dBZ & 0.420 [0.411, 0.431] & 0.329 [0.323, 0.337] \\
        & 30\,dBZ & 0.060 [0.050, 0.072] & 0.031 [0.026, 0.036] \\
        & 38\,dBZ & 0.029 [0.022, 0.036] & 0.012 [0.009, 0.015] \\
        \addlinespace
        \multirow{3}{*}{FAR}
        & 10\,dBZ & 0.727 [0.700, 0.756] & 0.767 [0.744, 0.792] \\
        & 30\,dBZ & 0.919 [0.900, 0.939] & 0.954 [0.944, 0.965] \\
        & 38\,dBZ & 0.958 [0.946, 0.972] & 0.982 [0.977, 0.988] \\
        \addlinespace
        \multirow{3}{*}{CSI}
        & 10\,dBZ & 0.198 [0.182, 0.213] & 0.158 [0.146, 0.169] \\
        & 30\,dBZ & 0.036 [0.028, 0.043] & 0.019 [0.015, 0.022] \\
        & 38\,dBZ & 0.017 [0.013, 0.022] & 0.007 [0.005, 0.009] \\
        \addlinespace
        \multirow{3}{*}{$\mathrm{FSS}_{128}$}
        & 10\,dBZ & 0.656 [0.628, 0.681] & 0.570 [0.546, 0.591] \\
        & 30\,dBZ & 0.273 [0.231, 0.321] & 0.176 [0.148, 0.208] \\
        & 38\,dBZ & 0.247 [0.188, 0.300] & 0.145 [0.110, 0.184] \\
        \addlinespace
        \multirow{3}{*}{$\mathrm{FSS}_{8}$}
        & 10\,dBZ & 0.422 [0.398, 0.444] & 0.350 [0.330, 0.368] \\
        & 30\,dBZ & 0.121 [0.100, 0.143] & 0.069 [0.058, 0.080] \\
        & 38\,dBZ & 0.077 [0.058, 0.093] & 0.036 [0.027, 0.043] \\
        \addlinespace
        \multirow{3}{*}{$\mathrm{FSS}_{32}$}
        & 10\,dBZ & 0.526 [0.500, 0.550] & 0.446 [0.424, 0.465] \\
        & 30\,dBZ & 0.188 [0.155, 0.223] & 0.114 [0.095, 0.135] \\
        & 38\,dBZ & 0.151 [0.112, 0.184] & 0.080 [0.059, 0.101] \\
        \addlinespace
        Pooled CRPS & --- & 12.26 [12.00, 12.53] & 11.23 [11.01, 11.44] \\
        \addlinespace
        CRPS & --- & 12.21 [12.04, 12.37] & 11.24 [11.05, 11.43] \\
        \bottomrule
    \end{tabular}
\end{table}

\subsection{Additional Case Studies}

This section presents additional case studies on the German and Italian domain in Supplementary Figs. \ref{fig:case-germany-convective} and \ref{fig:case-italy-light} together with their latest satellite inputs in Supplementary Figs. \ref{fig:case-germany-convective-inputs} and \ref{fig:case-italy-light-inputs} respectively.

\begin{figure}[htbp]
    \centering
    \includegraphics[width=\textwidth]{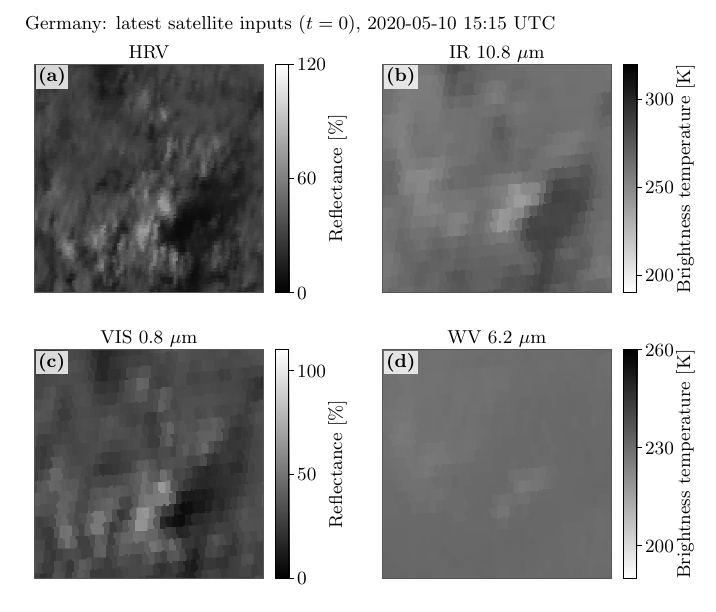}
    \caption{Latest satellite inputs ($t=0$) for the German convective case in Supplementary Fig.~\ref{fig:case-germany-convective}: \textbf{a}, high-resolution visible (HRV); \textbf{b}, 10.8-$\mu$m infrared; \textbf{c}, 0.8-$\mu$m visible; and \textbf{d}, 6.2-$\mu$m water vapour. Visible channels show reflectance and thermal channels brightness temperature.}
    \label{fig:case-germany-convective-inputs}
\end{figure}

\begin{figure}[htbp]
    \centering
    \includegraphics[width=\textwidth]{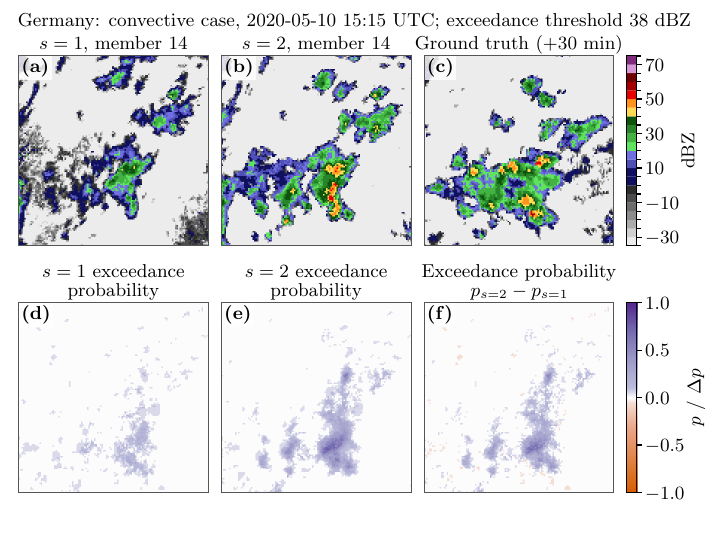}
    \caption{German convective case on 10 May 2020, 15:15 UTC and a lead time of 30 minutes. Each panel shows the same $128\times128$-pixel crop at $\SI{1}{\km}$ spatial resolution. \textbf{a,b}, The same paired ensemble member at guidance scales $s=1$ and $s=2$; \textbf{c}, ground truth. \textbf{d,e}, Exceedance probabilities $p_{s=1}$ and $p_{s=2}$, defined at each pixel as the fraction of the 20 ensemble members reaching or exceeding 38\,dBZ at the displayed lead time. \textbf{f}, The pixelwise probability difference $\Delta p=p_{s=2}-p_{s=1}$. Aggregated over all members, lead times, and pixels in this case, increasing the guidance scale from $s=1$ to $s=2$ raises CSI from 0.060 to 0.132 and the forecast exceedance fraction from 0.66\% to 2.28\%, compared with 1.36\% in the observations. The median member maximum increases from 54.08 to 59.71\,dBZ.}
    \label{fig:case-germany-convective}
\end{figure}

\begin{figure}[htbp]
    \centering
    \includegraphics[width=\textwidth]{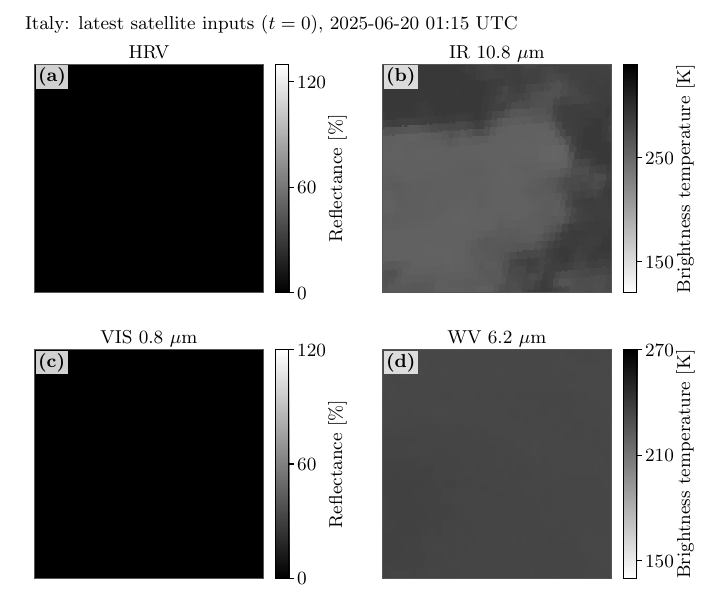}
    \caption{Latest satellite inputs ($t=0$) for the Italian light-rain case in Supplementary Fig.~\ref{fig:case-italy-light}. Panels and units are defined as in Supplementary Fig.~\ref{fig:case-germany-convective-inputs}; the near-zero visible-channel reflectances are expected at 01:15 UTC.}
    \label{fig:case-italy-light-inputs}
\end{figure}

\begin{figure}[htbp]
    \centering
    \includegraphics[width=\textwidth]{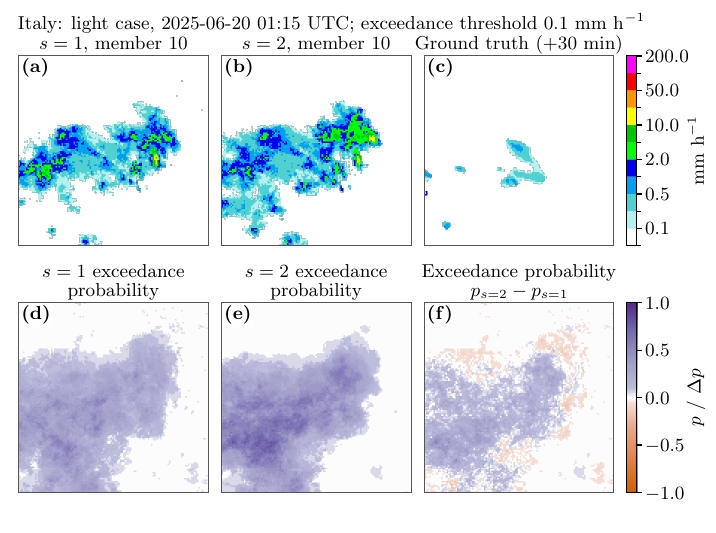}
    \caption{Italian light-rain case on 20 June 2025, 01:15 UTC and a lead time of 30 minutes. Crop dimensions, spatial resolution, panel definitions, and guidance scales are as in Supplementary Fig.~\ref{fig:case-germany-convective}, using an exceedance threshold of 0.10\,mm\,h$^{-1}$. Aggregated over all members, lead times, and pixels in this case, the observed exceedance fraction is 2.86\%, whereas the forecast fraction increases from 12.87\% at $s=1$ to 17.88\% at $s=2$. CSI rises only from 0.054 to 0.064, so the additional 5.01 percentage points of forecast event area primarily amplify false light-rain coverage.}
    \label{fig:case-italy-light}
\end{figure}

\clearpage
\putbib[references]

\end{bibunit}

\end{document}